%% file: iclr2027_conference.tex
\documentclass{article} %
\usepackage{iclr2027_conference,times}

\input{math_commands.tex}

\input{macro}

\usepackage[dvipsnames,HTML]{xcolor} 
\usepackage{hyperref}
\usepackage{url}
\newcommand\Mark[1]{\textsuperscript{#1}}

\title{\ourmethod: Second-order LLM Training at Linear Memory Cost via Nyström Sketching
}

\author{Beheshteh T. Rakhshan\Mark{$*$,1}, Sahar Rajabi\Mark{$*$,2}, Maziar Sargordi\Mark{3},\\
\textbf{Shikai Fang\Mark{4}, Guillaume Rabusseau\Mark{1,5}, Sirisha Rambhatla\Mark{2}}\\
\Mark{1}Mila \& DIRO, Universit\'e de Montr\'eal\\
\Mark{2}Critical ML, University of Waterloo\\
\Mark{3}Independent Researcher\\
\Mark{4}Zhejiang University\\
\Mark{5}Canada CIFAR AI Chair
}

\iclrfinalcopy %
\begin{document}

\maketitle
\def\thefootnote{$*$}\footnotetext{Equal contribution.}
\begin{abstract}
Training large language models (LLMs) entails a fundamental trade-off: memory-efficient optimizers such as Adam discard cross-parameter curvature, whereas full-curvature methods such as SOAP can accelerate convergence at prohibitive memory costs.
We introduce \textbf{\ourmethod}, a memory-efficient and full-curvature optimizer designed to resolve this bottleneck. \ourmethod leverages the randomized Nyström method to accurately approximate the left and right preconditioners in SOAP, and to reduce the optimizer's memory complexity from quadratic to linear in terms of model dimensions.
We subsequently reintegrate the off-subspace components to capture curvature information beyond the low-rank approximation, preserving rich curvature at minimal memory cost. 
We further propose \textbf{\ourqmethod}, a low-precision variant that aggressively compresses optimizer states. \ourqmethod reduces optimizer memory consumption by \textbf{over 50\%} compared to Muon when pre-training a LLaMA-1.3B architecture, all while maintaining strong and competitive predictive performance. Notably, \ourmethod operates with a smaller optimizer-state footprint than standard AdamW while reaching AdamW's final performance \textbf{26\% faster} in wall-clock time. Furthermore, our methods uniquely enable the pre-training of a 13B-parameter model on a single 80GB GPU, providing a scalable, efficient, and accessible approach to large-scale model optimization.
\end{abstract}

\input{sections/introduction}
\input{sections/background}
\input{sections/methodology}

\input{sections/theories}
\input{sections/related_works}

\input{sections/experiments}
\input{sections/conclusion}
\section*{Acknowledgment}
Sirisha Rambhatla and Sahar Rajabi acknowledge support from the Natural Sciences and Engineering Research Council of Canada (NSERC) Discovery Grant RGPIN-2022-03512 and the Val O’Donovan Chair endowment. Guillaume Rabusseau and Beheshteh T. Rakhshan acknowledge support from the CIFAR AI Chair Program. This work was made possible in part through computing resources provided by the Digital Research Alliance of Canada and Mila (mila.quebec).
\section*{AI Use Statement}
In this work, we used generative AI tools to provide feedback on our research methodology and to assist in implementing methods. We did not use generative AI for other tasks requiring disclosure, such as generating synthetic datasets, help develop theoretical models or conceptual frameworks, propose or refine hypothesis, design research methodology, assist in the writing of proofs,  provide critical ingredients for proving mathematical claims, assist with translation, clean and reformat dataset, support qualitative and thematic data analysis, developing theoretical models, formulating or proving mathematical claims, or interpreting the final results. 

Additionally, we utilized generative AI for several tasks with recommended disclosure: analyzing existing literature, creating scientific figures and images, and revising the paper draft to improve wording, flow, and overall readability. 

We have thoroughly reviewed and verified all AI-assisted work. We take full responsibility for the final content of this work, including all text, claims, code, and visual artifacts produced with the aid of generative AI.

\bibliography{bilbio}
\bibliographystyle{iclr2027_conference}

\appendix
\input{sections/appendix}

\end{document}

%% file: math_commands.tex
\usepackage{amsmath,amsfonts,bm}

\def\1{\bm{1}}

\DeclareMathAlphabet{\mathsfit}{\encodingdefault}{\sfdefault}{m}{sl}
\SetMathAlphabet{\mathsfit}{bold}{\encodingdefault}{\sfdefault}{bx}{n}

\DeclareMathOperator*{\argmin}{arg\,min}

%% file: macro.tex
\usepackage{fancyhdr}

\usepackage{amsmath}
\usepackage{amsthm}
\usepackage{amssymb}
\usepackage{graphicx}
\usepackage{accents}
\usepackage{hyperref}
\usepackage{bbm}
\usepackage{bm}
\usepackage{xspace}
\usepackage{amsmath}
\usepackage{wrapfig}

\usepackage{float}
\usepackage{scalerel}
\usepackage{tikz,amsmath,siunitx}
\usetikzlibrary{decorations.pathreplacing}
\usetikzlibrary{shapes}

\usepackage{fancybox}

\usetikzlibrary{positioning}
\usetikzlibrary{shapes.multipart,chains,}

\newcommand{\round}[1]{\left\lfloor #1 \right\rceil}

\newcommand{\RR}{\mathbb{R}}
\newcommand{\EE}{\mathbb{E}}

\newcommand{\ourmethod}{{Clean}\xspace}
\newcommand{\ourqmethod}{{Q-Clean}\xspace}

\definecolor{cMain}{RGB}{31,119,180}   %
\definecolor{cA}{RGB}{214,110,20}      %
\definecolor{cB}{RGB}{44,150,44}       %
\definecolor{cC}{RGB}{120,120,120}     %
\newcommand{\bM}[1]{\textcolor{cMain}{#1}}
\newcommand{\bA}[1]{\textcolor{cA}{#1}}
\newcommand{\bB}[1]{\textcolor{cB}{#1}}
\newcommand{\bC}[1]{\textcolor{cC}{#1}}

\usepackage{tikz,amsmath,siunitx}
\usetikzlibrary{decorations.pathreplacing}
\usetikzlibrary{shapes.geometric}
\usepackage{pifont}%

\newcommand{\ts}{\mathsf{T}}
\usepackage{titlesec}
\titleformat*{\subsubsection}{\normalfont}
\usepackage{multirow}
\usepackage{wrapfig}
\usepackage{etoolbox}
\usepackage{float}
\usepackage[T1]{fontenc}    %
\usepackage{hyperref}       %
\usepackage{url}            %
\usepackage{booktabs}       %
\usepackage{amsfonts}       %
\usepackage{nicefrac}       %
\usepackage{microtype}      %
\usepackage{xcolor}         %
\usepackage{algorithm}
\usepackage[noend]{algorithmic}
\usepackage[inline]{enumitem}
\usepackage{comment}
\usepackage{booktabs}
\usepackage{multirow}
\usepackage{caption}
\usepackage{subcaption}
\usepackage{booktabs}
\usepackage{multirow}
\usepackage{xcolor}
\usepackage{colortbl}                    %
\definecolor{ourcolor}{HTML}{EEF2FB}     %
\definecolor{ourqcolor}{HTML}{C1CBF8}

%% file: sections/introduction.tex
\section{Introduction}
\input{images/mem-simulation}
\vspace{-0.35cm}
Training large language models~(LLMs) forces a trade-off between memory and curvature. Adaptive optimizers store moment statistics that match or exceed the size of the model itself, and second-order methods such as Shampoo~\citep{gupta2018shampoo} and SOAP~\citep{vyas2024soap} store even more. The remedies that reduce this footprint typically discard curvature information, producing poorly conditioned updates and slowing convergence~\citep{gupta2018shampoo}. Balancing memory efficiency with curvature fidelity has therefore emerged as a central challenge for efficient LLM training.

A range of optimizers have been proposed, each making a different trade-off between memory and curvature. 
\emph{First–order methods} such as Adam \citep{kingma2014adam} and Adafactor \citep{shazeer2018adafactor} are lightweight optimizers that store only diagonal statistics. 
Their low memory cost makes them practical for large models.
\emph{Full–matrix second–order methods} such as Shampoo \citep{gupta2018shampoo} and KFAC \citep{martens2015optimizing} explicitly capture curvature and can accelerate training, but their quadratic state and computation make them prohibitive at LLMs scale. 
More recently, SOAP~\citep{vyas2024soap} achieved highly stable and full-rank preconditioning by applying Adam to the gradient projected onto the eigenbasis of Shampoo's preconditioner; however, it still relies on explicitly accumulating full-rank covariance matrices.
Several efficient optimizers have been proposed, but all of them have at least one optimizer memory state that grows quadratically with respect to model dimensions (i.e., $\mathcal{O}(mn)$)~\citep{romashov2026twosided,jordan2024muon,gong2026towards,liu2025cosmos}, making them difficult to scale. To resolve this, we introduce \textbf{\ourmethod}, a curvature-aware optimizer that achieves SOAP-level performance while its memory states scale linearly with respect to model dimensions. 

\ourmethod replaces explicit gradient covariance accumulations with randomized Nyström approximations, extracting highly accurate, compact estimations for the left and right Shampoo preconditioners. It then splits the update into four orthogonal blocks, $\texttt{Main}$, $\texttt{A}$, $\texttt{B}$, and $\texttt{C}$, and applies full Adam to \texttt{Main}. Rather than discarding gradients that fall outside the low-rank core, it applies pooled Adam-like updates to the three remaining blocks. Finally, every block is projected back to the parameter space, and \ourmethod recombines all four to produce a rich and full-rank update, as shown in Fig.~\ref{fig:clean}. We also introduce \textbf{\ourqmethod}, a quantized variant that aggressively compresses the stored optimizer states. 

To the best of our knowledge, \ourmethod is the first second-order SOAP-like optimizer whose memory grows only linearly with layer dimensions, requiring $\mathcal{O}((m+n)r)$ instead of $\mathcal{O}(mn)$ for a layer of size $m\times n$.
Extensive pre-training evaluations on LLaMA-based architectures demonstrate that our approach successfully retains the benefits of full-rank second-order preconditioning at a fraction of the optimizer memory cost. \ourmethod delivers validation perplexities highly competitive with SOAP, yet operates with a smaller memory footprint than AdamW, while showing significant speedup in comparison to Adam's convergence rate. Fig.~\ref{fig:mem-simulation} shows that across all the baselines, \ourmethod and \ourqmethod are the only optimizers that allow training a 13B model on an H100 GPU with 80GB of memory.

\input{images/clean}
In summary, \ourmethod is the first optimizer to perform SOAP-style preconditioning in the eigenbasis of randomized Nyström approximations of both Kronecker factors. The Nyström approximations allow the covariance factors to be updated without ever forming them explicitly, reducing
their memory cost from quadratic to linear in the layer dimensions for a fixed rank $r$. Even when the rank increases with the model dimensions, \ourmethod reduces the per-parameter cost as the model scales. Finally, we present \ourqmethod, a low-precision extension that quantizes optimizer states to achieve competitive performance while further reducing optimizer memory requirements.

%% file: images/mem-simulation.tex
\begin{wrapfigure}[17]{r}{0.5\textwidth}   %
  \vspace{-20pt}
    \centering
\includegraphics[width=\linewidth]{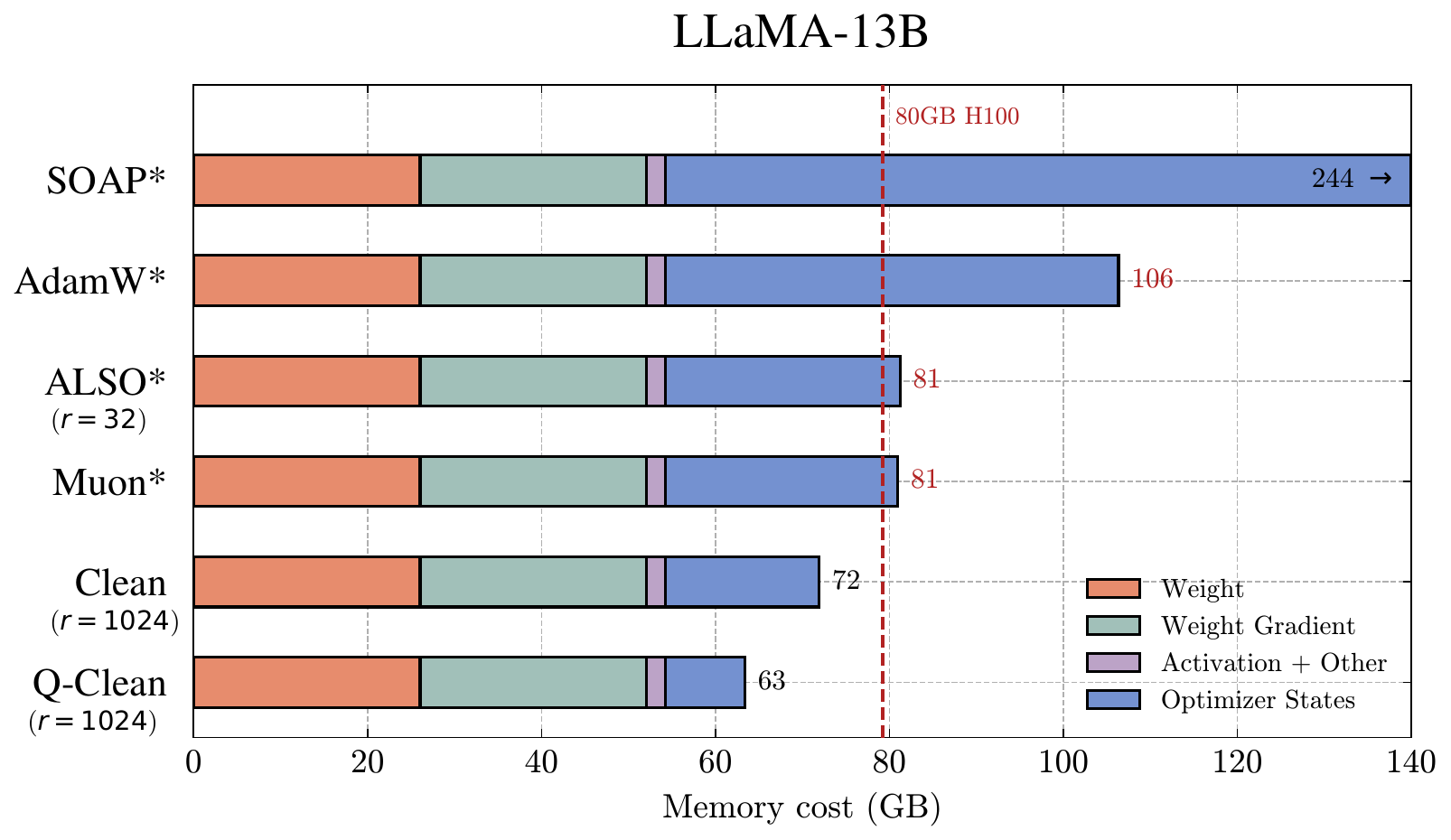}
    \caption{\small Measured memory consumption for pre-training a LLaMA-13B model on a single 80\,GB H100 GPU~(maximum sequence length of 1024, batch size of 1). All non-quantized matrices are stored in bfloat16. \emph{Activation + Other} combines activation memory with an estimated optimizer workspace. Methods marked $^{*}$ ran out of memory, so their optimizer state is estimated.}
    \label{fig:mem-simulation}
    \vspace{-20pt}
\end{wrapfigure}

%% file: images/clean.tex
\begin{figure}[t]
\begin{center}
\includegraphics[width=\linewidth]{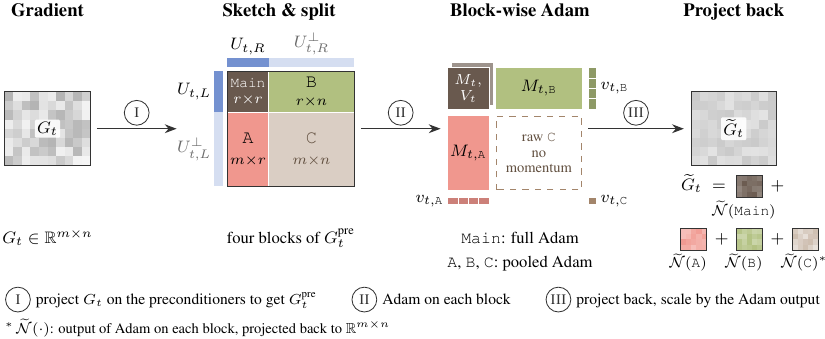}
\caption{\small \textbf{\ourmethod.} \textbf{(I)}~$G_t$ is projected via the estimated left and right preconditioners to yield $G_t^{\text{pre}}$, which is then partitioned into four blocks \texttt{Main}, $\texttt{A}$,
$\texttt{B}$, and $\texttt{C}$. \textbf{(II)}~Adam-style update is applied to each block, maintaining either full or pooled moments depending on the block's dimensions (only the shaded buffers are stored). \textbf{(III)}~The Adam output from each block is projected back into the parameter space and summed to form a full-rank update.}
\label{fig:clean}
\end{center}
\vspace{-20pt}
\end{figure}

%% file: sections/background.tex
\section{Background and Preliminaries}
Most deep learning optimizers can be viewed as preconditioned gradient methods.
Consider a weight matrix $W_t \in \RR^{m \times n}$ with vectorization
$w_t = \mathrm{vec}(W_t) \in \RR^{mn}$, loss function $\phi(\cdot)$ and stochastic gradient
$g_t = \nabla \phi_t(w_t) \in \RR^{mn}$. A generic preconditioned update takes the form
\begin{equation}\label{eq:generic-update}
\small
    w_{t+1} = w_t - \eta \, P_t^{-1} g_t,
\end{equation}
where $\eta > 0$ is the learning rate and $P_t \in \RR^{mn \times mn}$ is a
positive semi-definite~(PSD) preconditioner that encodes curvature information.
Different optimizers correspond to different choices of $P_t$, balancing the accuracy of the curvature with memory and computational costs.
We first review several widely used optimizers, and then introduce the technical tools on which our method builds.
\subsection{Optimizers as Preconditioners}
Eqn.~\eqref{eq:generic-update} provides a common lens for comparing optimizers: each corresponds to a particular choice of $P_t$, which dictates the balance between retained curvature information and memory requirements.
\paragraph{Adaptive optimization as preconditioning.} 
Full-matrix AdaGrad \citep{duchi2011adaptive} sets $P_t=\textstyle\sum_{s=1}^t g_s g_s^\top$, yielding well-conditioned updates but requiring $\mathcal{O}(m^2n^2)$ memory, which is infeasible for modern models with billions of parameters. Common approaches simplify $P_t$ to make optimization practical. \emph{Diagonal preconditioners}, used in optimizers such as Adam~\citep{kingma2014adam}, restrict $P_t$ to a diagonal matrix, capturing individual parameter scales but ignoring cross-parameter correlations, decreasing the optimizer memory complexity to $\mathcal{O}(mn)$.

\paragraph{Balanced preconditioning for matrix layers.}
A practical surrogate to approximate $P_t\in\RR^{mn\times mn}$ is maintaining left and right covariance accumulators:
\begin{equation}
\small
L_t = \sum_{s=1}^t G_s G_s^\top, \quad
R_t = \sum_{s=1}^t G_s^\top G_s,
\end{equation}
where $G_t$ is the gradient of $W$ at step $t$. 
The preconditioned gradient is then
\begin{equation}
\small
\tilde G_t = L_t^{-1/4} \, G_t \, R_t^{-1/4},
\end{equation}
 which captures the row and column curvature without ever forming the full preconditioner~\citep{gupta2018shampoo}. Concretely, $P_t$ is approximated by
the Kronecker product of two smaller factors,
$L_t$ and $R_t$, where the left preconditioner
$L_t \in \RR^{m \times m}$ acts on the rows of $G_t$, while $R_t \in \RR^{n \times n}$ acts on its columns.
Shampoo~\citep{gupta2018shampoo}, a popular second-order optimizer,
takes exactly this route and has become the standard Kronecker-factored
approximation to full-matrix AdaGrad~\citep{duchi2011adaptive}, which computes and stores per-layer $L_t$ and $R_t$, utilizing fractional powers $L_t^{-\alpha}, R_t^{-\alpha}$ as left and right preconditioners. The original $\alpha=1/4$ matches AdaGrad in theory, but \cite{anil2020scalable} found that $\alpha=1/2$ often improves stability. 
However, fully storing $L_t$ and $R_t$ still incurs quadratic memory in the layer width, which is prohibitive for wide layers.
\vspace{-0.1cm}
\paragraph{SOAP and eigenspace optimization.} 
Rather than explicitly computing the costly inverse fractional powers of $L_t$ and $R_t$ at every step, SOAP~\citep{vyas2024soap} periodically computes their eigen decompositions. 
It then projects the gradients into this slowly changing principal curvature eigenspace, applies a coordinate-wise Adam update, and projects the step back to the original parameter space. 
Ultimately, this hybrid approach unifies the structural curvature benefits of second-order optimization with the low per-step cost of diagonal methods, empirically reducing both the number of iterations and the wall-clock time compared to standard Shampoo and AdamW. However, SOAP's memory footprint remains $\mathcal{O}(m^2 + n^2)$, the same
as Shampoo's, since it must still store the full $m \times m$ and
$n \times n$ Kronecker factors, $L_t$ and $R_t$ at each step. The SOAP algorithm is provided in Appendix~\ref{app:soap}.

\subsection{Randomized Nyström preconditioner} 
The randomized Nyström approximation~\citep{gittens2016revisiting} provides a framework for efficiently representing symmetric PSD matrices using low-rank approximations. Given a symmetric PSD matrix $A\in\RR^{n\times n}$ and a random~(Gaussian) matrix $\Omega\in\mathbb{R}^{n\times r}$
where $r$ is a rank parameter, the randomized Nyström approximation is given by:
\begin{equation}\label{eqn:nystrom}
\small
    Y = A\Omega, \quad B = \Omega^\top A \Omega, \quad \tilde A = Y B^\dagger Y^\top \simeq A, 
\end{equation}
where $\tilde{A}$ is a rank-$r$ surrogate, requiring only
$\mathcal{O}(nr)$ storage instead of $\mathcal{O}(n^2)$ when
$r\ll n$.
Strong theoretical guarantees \citep{gittens2016revisiting} and practical efficiency of Nyström approximations, make them attractive for optimizer states.
However, a naive implementation based on Eqn.~\eqref{eqn:nystrom} is numerically unstable~\citep{frangella2023randomized}. Alg.~\ref{alg:nystrom} in Appendix~\ref{app:nystrom} presents a numerically stable approach for constructing a Nyström approximation~\citep{tropp2017fixed}. 
Given the eigenvalue decomposition $\tilde{A} = U \Lambda U^\top$ returned by
Alg.~\ref{alg:nystrom}, where $U \in \RR^{n\times r}$ has
orthonormal columns and
$\Lambda = \mathrm{diag}(\lambda_1, \dots, \lambda_r)$ with
$\lambda_1 \ge \dots \ge \lambda_r \ge 0$, the Nyström preconditioner is obtained by:
\begin{equation}\label{eq:nystrom-precond}
\small
    P = (\lambda_r + \mu)^{-1}\, U (\Lambda + \mu I) U^\top + (I - U U^\top).
\end{equation}
In effect, $P^{-1}$ scales down each of the top-$r$ eigendirections of $A$ by the factor $(\lambda_r+\mu)/(\lambda_i+\mu) \le 1$, so that directions with larger eigenvalues are scaled down more, while acting as the identity on $U^\perp$.

%% file: sections/methodology.tex
\section{Methodology}
In this section, we present our second-order memory‑efficient optimizer, \textbf{\ourmethod}, starting by approximating the left and right preconditioners in SOAP via the Nyström method.
During training, the singular values of the gradients decay rapidly, concentrating most of their energy in a small subspace~\citep{zhao2024galore}. Since the eigenvalues of $G_tG_t^\top$ and $G_t^\top G_t$ are the squared singular values of $G_t$, these covariances exhibit an even sharper spectral decay, making the SOAP factors $L_t$ and $R_t$ natural targets for low-rank approximation. \ourmethod exploits this structure to reduce SOAP's $\mathcal{O}(m^2 + n^2)$ memory to $\mathcal{O}((m+n)r)$, which is linear in $m$ and $n$.

\paragraph{Smoothed covariances.}
Similar to SOAP, \ourmethod preconditions with exponential moving averages of the gradient covariances (Eqn.~\eqref{eq:smoothed-cov}).
However, it never stores these covariances in full, instead, it maintains
the low-rank approximations as in Eqn.~\eqref{eq:left-right-nys},
\begin{equation}\label{eq:smoothed-cov}
\small
  L_t = \beta_2\, L_{t-1} + (1-\beta_2)\, G_tG_t^\top,
  \quad
  R_t = \beta_2\, R_{t-1} + (1-\beta_2)\, G_t^\top G_t.
\end{equation}
The averaging prevents abrupt changes of basis across refreshes. For
$\beta_2 \in [0,1]$, each update is a convex combination of PSD matrices, so $L_t$ and $R_t$ remain PSD and therefore admit
randomized Nyström approximations~\citep{frangella2023randomized}. 

\paragraph{Low-rank covariance factors.}
\ourmethod never forms $L_t$ or $R_t$ explicitly. Instead, it maintains
rank-$r_1$ and rank-$r_2$ Nyström approximations
\begin{align}\label{eq:left-right-nys}
\small
  L_t \approx U_{t,L}\Lambda_{t,L}U_{t,L}^\top,
  \qquad
  R_t \approx U_{t,R}\Lambda_{t,R}U_{t,R}^\top,
\end{align}
where $U_{t,L} \in \RR^{m\times r_1}$ and
$U_{t,R} \in \RR^{n\times r_2}$ have orthonormal columns,
$\Lambda_{t,L} \in \RR^{r_1\times r_1}$ and $\Lambda_{t,R} \in \RR^{r_2\times r_2}$, are diagonal and non-negative,
and $r_1,r_2 \ll \min(m,n)$. \ourmethod only stores thin factors $U_{t,L}, \Lambda_{t,L}$ and $U_{t,R}, \Lambda_{t,R}$. To update them at step $t$, we replace $L_{t-1}$ in the
Eqn.~\eqref{eq:smoothed-cov} with its approximation
and access the result only through a random sketch. Let
$\Omega_L \in \RR^{m\times (r_1+p)}$ and $\Omega_R \in \RR^{n\times (r_2+p)}$
be Gaussian random matrices, where $p$ is an oversampling
parameter~\citep{halko2011finding}. The left sketch is
\begin{equation}\label{eq:sketched-preconditioner}
\small
    L_t\Omega_L \approx \beta_2\, U_{t-1,L}\Lambda_{t-1,L}\big(U_{t-1,L}^\top\Omega_L\big)
               + (1-\beta_2)\, G_t\big(G_t^\top\Omega_L\big),
\end{equation}
and $R_t\Omega_R$ is computed analogously (see Appendix~\ref{app:nystrom} for details of Nyström approximations). Both require only thin matrix
products with the stored factors and gradient, so no
$m\times m$ or $n\times n$ matrix is ever formed. Applying
Alg.~\ref{alg:warmup-nystrom} in Appendix~\ref{app:nystrom} to these sketches yields the updated
factors in Eqn.~\eqref{eq:left-right-nys}.

Without loss of generality, we assume
$r_1 = r_2 = r$. SOAP accumulates $L_t$ and $R_t$ at every step, whereas \ourmethod
accumulates them only every $T$ steps. Moreover, it only stores the low-rank factors, at a cost of $(m + n + 2)r$
parameters, entirely eliminating the quadratic dependence of memory on the
layer dimensions. As in SOAP, the factors are refreshed
every $T$ iterations and reused in between, so $L_{t-1}$ and
$R_{t-1}$ always refer to the most recent refresh.

\paragraph{Block-wise low-rank preconditioning.}
Following~\citep{gupta2018shampoo,morwani2024new}, we precondition the gradient as
$G_t^{\mathrm{pre}} = L_t^{-1/2} G_t R_t^{-1/2}$. To expand $G_t^{\mathrm{pre}}$ with the Nyström
preconditioner in Eqn.~\eqref{eq:nystrom-precond}, for $S \in \{L, R\}$, we write
\begin{equation}
\small
  P_S = U_{t,S}U_{t,S}^\top, \qquad
  P_S^{\perp} = I - P_S, \qquad
  D_S = (\Lambda_{t,S} + \mu I)^{-1/2}, \qquad
  c_S = (\lambda_{S,r} + \mu)^{1/2},
\end{equation}
where $\lambda_{S,r}$ is the smallest retained eigenvalue. This gives
\begin{align}\label{eq:G_pre_decomposition}
\small
  G_t^{\mathrm{pre}}
  &= \underbrace{c_L c_R\, U_{t,L} D_L\,
       \textcolor{cMain}{U_{t,L}^\top G_t U_{t,R}}\,
       D_R U_{t,R}^\top}_{\texttt{Main}}
   \;+\; \underbrace{c_R\,
       \textcolor{cA}{P_L^{\perp} G_t U_{t,R}}\,
       D_R U_{t,R}^\top}_{\texttt{A}} \nonumber\\[4pt]
  &\quad
   +\; \underbrace{c_L\, U_{t,L} D_L\,
       \textcolor{cB}{U_{t,L}^\top G_t P_R^{\perp}}}_{\texttt{B}}
   \;+\; \underbrace{\textcolor{cC}{P_L^{\perp} G_t P_R^{\perp}}}_{\texttt{C}}.
\end{align}
Running Adam directly on $G_t^{\mathrm{pre}}$ would require $\mathcal{O}(mn)$ memory for its
moments. \ourmethod{} instead keeps separate moment statistics for each block
of Eqn.~\eqref{eq:G_pre_decomposition} according to its size. For the \texttt{Main} block, Adam runs on the
two-sided projection $U_{t,L}^\top G_t U_{t,R} \in \mathbb{R}^{r\times r}$. In addition, the cross
components in blocks $\texttt{A}$ and $\texttt{B}$ lie outside one subspace but inside the other, so
they are also low-dimensional:
\begin{equation}
\small
  G_{t,\texttt{A}} = P_L^{\perp} G_t U_{t,R} \in \mathbb{R}^{m\times r},
  \qquad
  G_{t,\texttt{B}} = U_{t,L}^\top G_t P_R^{\perp}\in \mathbb{R}^{r\times n}.
\end{equation}
Only the component in block~\texttt{C} is orthogonal to both subspaces and is a full $m\times n$ matrix.
Rather than discarding the components outside the \texttt{Main} block, \ourmethod applies
Adam-style updates to each block with its own moment estimates. However, applying Adam to these blocks as in \cite{romashov2026twosided}, requires buffers of size $\mathcal{O}(r^2)$, $\mathcal{O}(mr)$, $\mathcal{O}(rn)$, and
$\mathcal{O}(mn)$ for \texttt{Main}, \texttt{A}, \texttt{B}, and \texttt{C}, respectively. The
first three are low-rank, but the last is as large as running Adam directly on
$G_t^{\mathrm{pre}}$, reintroducing exactly the cost that \ourmethod aims to remove.

\input{images/grad_frac}
Fig.~\ref{fig:block-share} illustrates the fraction of the gradient norm, averaged across all 24 instances of a layer type in a LLaMA-350M, for each of the blocks in Eqn.~\eqref{eq:G_pre_decomposition}. \texttt{Main} dominates across nearly all layers, while \texttt{C} consistently contributes less. 
The figure also underscores the necessity of reintegrating information from blocks other than the \texttt{Main}, as they carry a substantial portion of the training signal, particularly in later iterations.
We therefore tailor the moment state to each
block, as summarized in Table~\ref{tab:block_state_size}.
\input{tables/block-sizes}

For the $\texttt{Main}$ block, we apply regular Adam, maintaining entry-wise first and second moments ($M$ and $V$), which each require $\mathcal{O}(r^2)$ memory. 
\begin{equation}
\small
    M_t \gets \beta_1 M_{t-1} + (1-\beta_1)S_t, \quad
    V_t \gets \beta_2 V_{t-1} + (1-\beta_2)S_t^{\odot 2},
\end{equation}
where $\odot$ denotes the Hadamard product. For blocks $\texttt{A}$ and $\texttt{B}$, we retain the full-resolution first moments but spatially pool the second moments. Specifically, by assuming a shared scale for each column in $\texttt{A}$ and each row in $\texttt{B}$, we reduce the second moment to a single scalar per column or row. 
\begin{align}
\small
    \begin{split}
        & M_{t,\texttt{A}} \gets \beta_1 M_{t-1,\texttt{A}} + (1-\beta_1)G_{t,\texttt{A}}, \quad
        v_{t,\texttt{A}} \gets \beta_2 v_{t-1,\texttt{A}} + (1-\beta_2)\,\mathbf{1}^\top (G_{t,\texttt{A}}^{\odot 2}) \\
        & M_{t,\texttt{B}} \gets \beta_1 M_{t-1,\texttt{B}} + (1-\beta_1)G_{t,\texttt{B}}, \quad
        v_{t,\texttt{B}} \gets \beta_2 v_{t-1,\texttt{B}} + (1-\beta_2)\,(G_{t,\texttt{B}}^{\odot 2})\mathbf{1}
    \end{split}
\end{align}
Finally, for block $\texttt{C}$, we omit the first moment entirely, utilizing the raw gradient instead, and employ its squared Frobenius norm as a scalar second moment. 

\paragraph{Projection-aware adaptive moments.}
On subspace-refresh steps, when the bases $U_{t,S}$ change ($S\in \{L, R\}$), the moments will get misaligned. The first moments $M$, $M_\texttt{A}$, and $M_\texttt{B}$ are valid operators in the original coordinates, and they can be moved exactly into the new coordinates by:
\begin{align}
\label{eq:rotate-moments}
\small
\begin{split}
    & M \leftarrow (U_{t,L}^\top U_{t-1, L}) M (U_{t-1,R}^\top U_{t, R}), \\
     M_{\texttt{A}} \leftarrow U_{t,L}^\perp &M_{\texttt{A}}(U_{t-1,R}^\top U_{t, R}), \qquad M_{\texttt{B}} \leftarrow (U_{t,L}^\top U_{t-1, L}) M_{\texttt{B}}U_{t,R}^\perp,
\end{split}
\end{align}
However, the second moments cannot be exactly transformed. $V$ is an exponential moving average of element-wise squared gradients, tracking only the diagonal (per-coordinate) variances. Since element-wise squaring does not commute with basis rotations, calculating the true rotated variances would require unrecorded off-diagonal cross-terms. 
Similar to SOAP \citep{vyas2024soap}, we explicitly rotate the first moment $M$, and leave $V$ in place. By relying on warm-start refreshes (see Appendix~\ref{app:nystrom}), we constrain the overall frame drift, ensuring the stale-basis error in $V$ remains sufficiently small.

For \texttt{A} and \texttt{B}, we transport the pooled second moments on change of basis. Since each value stores the squared gradient along one direction, we use the squared overlap between the old and new bases:
\begin{equation}
\label{eq:rotate-pooled}
\small
v_\texttt{A}^\top \leftarrow v_\texttt{A}^\top (U_{t-1,R}^\top U_{t,R})^{\odot 2},
\qquad
v_\texttt{B} \leftarrow (U_{t,L}^\top U_{t-1,L})^{\odot 2} v_\texttt{B}.
\end{equation}
Finally, the scalar $v_\texttt{C}$ does not need to be rotated because it has no direction associated with it.
\paragraph{Projection back and parameter update.} After applying the block-wise Adam-style updates, each block is projected back to the original parameter space to construct the final weight update, as shown in Eqn.~\eqref{eq:G_pre_decomposition}. Note that the momentum terms used in this step are bias-corrected, and similar to SOAP~\cite{vyas2024soap}, we are using Adam scaling factors while projecting back, which eliminates the need for $D_S$ and $c_S$ factors in reconstructing $G_t^{\text{pre}}$ (\ref{eq:G_pre_decomposition}).
\begin{align}\label{eq:back-projections}
\small
    \widetilde{G}_t 
    = 
    \underbrace{U_{t,L} \mathcal{N}(\hat{M_t}, \hat{V_t}) U_{t,R}^\top}_{\text{Block}~\texttt{Main}~\text{(Full Adam)}}
    + 
    \underbrace{\mathcal{N}(\hat{M}_{t,\texttt{A}}, \hat{v}_{t,\texttt{A}} )U_{t,R}^\top}_{\text{Block}~\texttt{A}~\text{(Pooled Adam)}}
    +
    \underbrace{U_{t,L} \mathcal{N}(\hat{M}_{t,\texttt{B}}, \hat{v}_{t,\texttt{B}})}_{\text{Block}~\texttt{B}~\text{(Pooled Adam)}}
    + 
    \underbrace{\mathcal{N}(\hat{C}_t, \hat{v}_{t,\texttt{C}})}_{\text{Block}~\texttt{C}~\text{(Pooled Adam)}},
\end{align}
where $\mathcal{N}(\hat{M}, \hat{V}) = \hat{M} / (\sqrt{\hat{V}} + \epsilon)$
for the corresponding bias-corrected moments $\hat{M}$ and $\hat{V}$.
The weight update is then
$
    W_t \leftarrow W_{t-1} - \alpha \eta \cdot \widetilde G_t
$
where $\eta$ is the learning rate and $\alpha$ is a scale factor.

\input{tables/mem_complexities} 
Clean avoids building any \(\mathcal{O}(mn)\) memory while leveraging full-rank information and maintaining performance by (i) representing preconditioners with Nyström factors \(U_{t,*},\Lambda_{t,*}\), (ii) operating on four complementary blocks, and (iii) updating the stored low-rank moments when bases change. Memory therefore grows only \emph{linear} with respect to model dimensions. Furthermore, our warm-start Nystr\"om variant (see Alg.~\ref{alg:warmup-nystrom} in Appendix~\ref{app:nystrom}) enables gradual basis evolution, improving stability during periodic subspace refreshes. As shown in Table~\ref{tab:optimizer_memory}, \ourmethod is the only evaluated optimizer that scales linearly in state memory.

\input{algorithms/clean}

\section{\ourqmethod: Low-Precision Second-Order Optimization}

To further reduce memory overhead, we explore the stability of \ourmethod under quantization, a technique that stores matrices as low-bit integers. In this section, we briefly introduce \ourqmethod, which employs quantized optimizer states. More details are included in Appendix~\ref{app:quantization}.

Following \citep{dettmers20218}, block-wise quantization partitions a flattened matrix $X$ into contiguous blocks of size $B$ and quantizes each block independently. With an $n$-bit asymmetric (affine) quantizer, block $i$ uses a scale
$s_i$ and a zero-point
$z_i$ to represent a value in lower precision.

Per-block scaling limits the effect of outliers to a single block rather than coarsening the quantization grid of the entire
matrix~\citep{dettmers2022gpt3}. However, because optimizer states are continually updated, round-to-nearest scheme discards any update smaller than half a quantization step, so small but consistent updates never accumulate. To remove this bias, the linear quantizer can instead use stochastic rounding~\citep{von1947numerical,gupta2015deep}, which is unbiased in
expectation, and therefore prevents small updates from vanishing over time.

For the first-moment and projection matrices, we apply the block-wise linear asymmetric scheme mentioned above. In contrast, the second-moments are strictly non-negative and exhibit a highly skewed distribution with a large dynamic range~(see Appendix~\ref{app:quantization}). Because linear quantization is suboptimal for such statistics, we instead adopt the non-linear quantization implementation from \texttt{bitsandbytes} \citep{dettmers20218}, which is also used in the Adam-8bit optimizer. We quantize the first moments of all blocks listed in Table~\ref{tab:block_state_size}, the second moment of the \texttt{Main}, and the projection matrices in INT8. The second moments of blocks \texttt{A} and \texttt{B} are kept in full precision: each is a vector of size $r$, so its memory footprint is negligible and quantizing it would yield no meaningful savings.

%% file: images/grad_frac.tex
\begin{figure}[t]
 \small
    \centering
    \begin{subfigure}{0.24\textwidth}
        \centering
        \includegraphics[width=\linewidth, trim=0 0 0 0, clip]{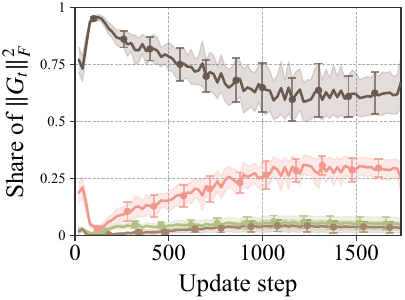}
        \caption{\scriptsize Attention-Query Proj.}
    \end{subfigure}
    \begin{subfigure}{0.24\textwidth}
        \centering
        \includegraphics[width=\linewidth, trim=0 0 0 0, clip]{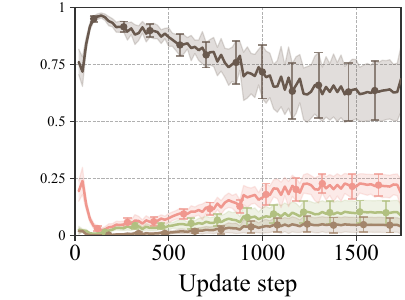}
        \caption{\scriptsize Attention-Key Proj.}
    \end{subfigure}
    \begin{subfigure}{0.24\textwidth}
        \centering
        \includegraphics[width=\linewidth, trim=0 0 0 0, clip]{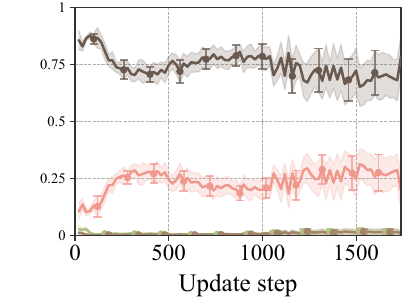}
        \caption{\scriptsize Attention-Value Proj.}
    \end{subfigure}
    \begin{subfigure}{0.24\textwidth}
        \centering
        \includegraphics[width=\linewidth, trim=0 0 0 0, clip]{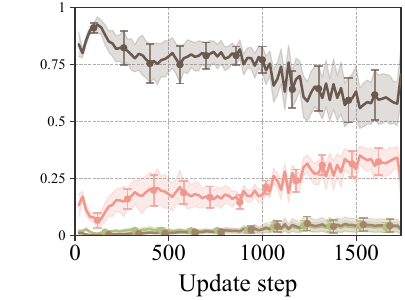}
        \caption{\scriptsize Attention-Output Proj.}
    \end{subfigure}

    \begin{subfigure}{0.24\textwidth}
        \centering
        \includegraphics[width=\linewidth, trim=0 0 0 0, clip]{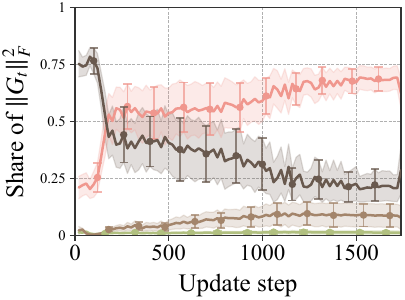}
        \caption{\scriptsize MLP-Gate Proj.}
    \end{subfigure}
    \begin{subfigure}{0.24\textwidth}
        \centering
        \includegraphics[width=\linewidth, trim=0 0 0 0, clip]{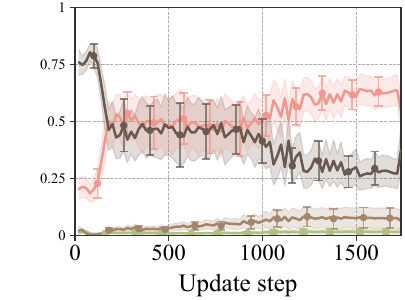}
        \caption{\scriptsize MLP-Up Proj.}
    \end{subfigure}
    \begin{subfigure}{0.24\textwidth}
        \centering
        \includegraphics[width=\linewidth, trim=0 0 0 0, clip]{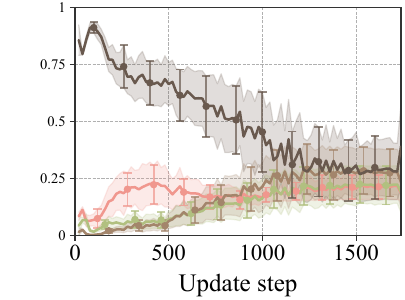}
        \caption{\scriptsize MLP-Down Proj.}
    \end{subfigure}
    \begin{subfigure}{0.24\textwidth}
        \centering
        \includegraphics[height=0.73\linewidth, trim=0 0 0 0, clip]{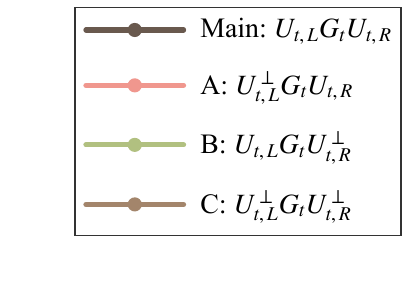}
        \caption{\scriptsize Legend}
    \end{subfigure}

    \caption{\small Fraction of the gradient norm captured in each of the four blocks. Generally, \texttt{Main} and \texttt{A} are dominant; the contributions of \texttt{B} and \texttt{C} slowly increase only towards the end of training, where the gradients become noisier. Values are averaged across all 24 instances of a layer type in a LLaMA-350M architecture.}
    \label{fig:block-share}
\end{figure}

%% file: tables/block-sizes.tex
\begin{table}[h]
\centering
\small
\caption{\textbf{Moments per block.} $\EE_{\beta}[\cdot]$ denotes an exponential moving
average with rate $\beta \in \{\beta_1, \beta_2\}$, and memory counts stored floats. \texttt{Main} keeps both full moments; $\texttt{A}$ and $\texttt{B}$ keep
full first moments, but sum their second moments over the
long dimension, leaving one value per column or row; $\texttt{C}$
keeps no first moment and a single scalar second moment, since per-entry states
would cost $\mathcal{O}(mn)$.}
\begin{tabular}{l|l|l|l|l}
\toprule
Block & Adam's Input & $1^{st}$ Moment $M$ & $2^{nd}$ Moment $V$ & Memory \\
\midrule
$\texttt{Main}$ & $S_t = U_{t,L}^\ts G_tU_{t,R} \in \RR^{r \times r}$ & $\EE_{\beta_1}[S_t]$
     & $\EE_{\beta_2}[S_t^{\odot 2}]$ & $2r^2$ \\[2pt]
$\texttt{A}$  & $G_{t,A}\in\RR^{m \times r}$ & $\EE_{\beta_1}[G_{t,A}]$
     & $\EE_{\beta_2}\big[\sum_i (G_{t,A})_{ij}^2\big]$ & $mr+r$ \\[2pt]
$\texttt{B}$  & $G_{t,B}\in\RR^{r \times n}$ & $\EE_{\beta_1}[G_{t,B}]$
     & $\EE_{\beta_2}\big[\sum_j (G_{t,B})_{ij}^2\big]$ & $rn+r$ \\[2pt]
$\texttt{C}$  & $C_t \in \RR^{m \times n}$ & none, raw $C_t$
     & $\EE_{\beta_2}\big[\|C_t\|_F^2\big]$ & $1$ \\
\bottomrule
\end{tabular}\label{tab:block_state_size}
\end{table}

%% file: tables/mem_complexities.tex
\begin{wraptable}{r}{0.5\linewidth}
\vspace{-12pt}
  \centering
  \caption{Optimizer-state memory requirement for an $m \times n$ matrix.}
  \label{tab:optimizer_memory}
  \resizebox{\linewidth}{!}{%
  \begin{tabular}{lr}
    \toprule
    Optimizer & State size \\
    \midrule
    Adam       & $2mn$ \\
    SOAP       & $2mn + 2m^2 + 2n^2$ \\
    Muon       & $mn$ \\
    ALSO       & $mn + 2mr + 2nr + ~~r^2 + 2r + O(1)$ \\
    \ourmethod~(Ours) & $2mr + 2nr + 2r^2 + 4r + O(1)$ \\
    \bottomrule
  \end{tabular}
  }
\vspace{-10pt}
\end{wraptable}

%% file: algorithms/clean.tex
\newcommand{\algphase}[1]{\item[] \textbf{#1}}
\newcommand{\cmt}[1]{\hfill\textcolor{gray}{$\triangleright$~#1}}
\begin{algorithm}[th]
\small
\caption{\ourmethod at iteration $t$. Colors mark the blocks
of Eqn.~\eqref{eq:G_pre_decomposition}: \bM{Main}, \bA{$A$}, \bB{$B$}, \bC{$C$}.
\\
Between refreshes, the previous
factors are reused.}
\label{alg:method}
\begin{algorithmic}[1]
\REQUIRE $G_t \in \RR^{m\times n}$, step size $\eta$, scale $\alpha$,
         $\beta_1, \beta_2, \epsilon$, rank $r$, refresh period $T$, and $t \ge 1$
\vspace{3pt}
\IF{$t-1 \bmod T = 0$}
\STATE $U_{t,L},\Lambda_{t,L}\gets\mathrm{Nys}\left(\beta_2 U_{t-1,L}\Lambda_{t-1,L}U_{t-1,L}^\top + (1-\beta_2)G_tG_t^\top\right)$
 \cmt{Alg.~\ref{alg:warmup-nystrom}}
\STATE $U_{t,R},\Lambda_{t,R}\gets\mathrm{Nys}\left(\beta_2 U_{t-1,R}\Lambda_{t-1,R}U_{t-1,R}^\top + (1-\beta_2)G_t^\top G_t\right)$
 \cmt{Alg.~\ref{alg:warmup-nystrom}}
  \STATE Rotate $M, M_A, M_B, v_A, v_B$ into the new bases
         \cmt{Eqn.~\eqref{eq:rotate-moments}~\& Eqn.~\eqref{eq:rotate-pooled}}
\ENDIF
    \STATE $U_{t,L},\Lambda_{t,L}\gets U_{{t-1},L},\Lambda_{{t-1},L}$ 
    \STATE $U_{t,R},\Lambda_{t,R}\gets U_{{t-1},R},\Lambda_{{t-1},R}$
    \\
   \hrulefill
\vspace{3pt}
\STATE $\bM{S_t \gets U_{t,L}^\ts G_t U_{t,R}}$, \;
       $\bA{G_{t,A} \gets P_L^{\perp} G_t U_{t,R}},$ \;
       $\bB{G_{t,B} \gets U_{t,L}^\ts G_t P_R^{\perp}}$, \;
       $\bC{C_t \gets P_L^{\perp} G_t P_R^{\perp}}$
       \cmt{Eqn.~\eqref{eq:G_pre_decomposition}}
\vspace{3pt}
\STATE $\bM{M_t}$, \;
       $\bM{V_t}$, \;
       $\bA{M_{t,A}}$, \;
       $\bA{v_{t,A}}$, \;
       $\bB{M_{t,B}}$, \;
       $\bB{v_{t,B}}$, \;
       $\bC{v_{t,C}}$
       \cmt{Moments, Table~\ref{tab:block_state_size}}
       \STATE $\widetilde G_t \gets
        \bM{U_{t,L}\,\mathcal{N}(M_t, V_t)\,U_{t,R}^\ts}
      + \bA{\mathcal{N}(M_{t,A}, v_{t,A})\,U_{t,R}^\ts}
      + \bB{U_{t,L}\,\mathcal{N}(M_{t,B}, v_{t,B})}
      + \bC{\mathcal{N}(C_t, v_{t,C})}$
       \cmt{Eqn.~\eqref{eq:back-projections}}
       \\
       \hrulefill
\vspace{3pt}

\STATE $W_t \gets W_{t-1} - \alpha\eta\,\widetilde G_t$
\end{algorithmic}
\end{algorithm}

%% file: sections/related_works.tex
\section{Related Work}
\label{sec:related}
\vspace{-0.2cm}
\paragraph{Low-rank Projection Methods.}
A popular approach to reducing optimizer memory is to project gradients onto a low-rank subspace. GaLore~\citep{zhao2024galore} introduced this strategy by performing gradient updates within a low-rank subspace defined by the singular value decomposition (SVD) of gradients. A family of works follows this line of optimizers~\citep{chen2410fira, zhu2025apollo, liang2024memory, he2024subspace}. While effective at saving memory, this one-sided projection can be unstable as the subspace drifts over time. LDAdam~\citep{robert2024ldadam} and Subtrack++~\citep{rajabi2025subtrack++} attempted to mitigate this by incorporating projection-aware updates and more robust subspace tracking mechanisms. However, these methods operate on one-sided approximations and do not capture the row-column geometry of the gradient covariances.
\vspace{-0.2cm}
\paragraph{Second-order Preconditioning.}
Another line of work focuses on capturing curvature information. K-FAC~\citep{martens2015optimizing} approximates the Fisher information matrix with Kronecker products, while Shampoo~\citep{gupta2018shampoo} directly approximates the full-matrix AdaGrad~\citep{duchi2011adaptive} preconditioner by maintaining separate row and column preconditioners. These balanced, two-sided preconditioners are powerful but require storing full-sized covariance matrices, making them memory- and compute-intensive. SOAP~\citep{vyas2024soap} combines Shampoo's preconditioning with first-order updates but still retains the costly full-matrix accumulators.
\vspace{-0.2cm}
\paragraph{Low-rank Second-order Methods.}
To achieve the benefits of second-order preconditioning without the prohibitive memory costs, recent methods have proposed efficient second-order optimizers. Alice~\citep{gong2026towards} provides a low-rank extension that converts general full-rank optimizers into memory-efficient approximations by relying on structured Fisher information matrix approximations. Building on this, ALSO~\citep{romashov2026twosided}, extends toward a two-sided approximation of SOAP~\citep{vyas2024soap}, while compensating for the discarded residual blocks outside the core low-rank projection. Another direction leverages the varying importance of eigensubspaces; for instance, COSMOS~\citep{liu2025cosmos} introduces a hybrid approach that applies memory-intensive SOAP updates exclusively to the leading eigensubspace of the gradient matrix, while utilizing a computationally cheaper, Muon-like update~\citep{jordan2024muon} for the residual subspace.
\vspace{-0.2cm}
\paragraph{Low Precision Training with Quantization.}
Quantization improves memory efficiency by representing data in reduced-precision formats. These techniques broadly fall into Post-Training Quantization~(PTQ), where pre-trained models are quantized without additional training~\citep{ashkboos2025halo, dettmers2023qlora, ashkboos2024quarot}, and Quantization-Aware Training~(QAT), which integrates quantization directly into the training loop. A major challenge in QAT is training instability. Inspired by~\citet{izmailov2018averaging}, SWALP~\citep{yang2019swalp} mitigates this using stochastic weight averaging, though it incurs significant memory overhead when applied to large foundation models. Q-Galore~\citep{zhang2024q} substantially reduces memory consumption by integrating full-model quantization with the low-rank gradient projection of GaLore~\citep{zhao2024galore}. Similarly, 4-bit Shampoo~\citep{wang20244} quantizes the left and right eigenvector matrices and applies Björck orthonormalization to restore orthogonality after quantization.

%% file: sections/experiments.tex
\section{Experiments}
\label{sec:experiment}
\input{images/val_perplexity}
\paragraph{Setup.}
We evaluate \ourmethod and \ourqmethod on pre-training LLaMA-based architectures~\citep{touvron2023llama} with 350M and 1.3B parameters on the C4 dataset~\citep{raffel2020exploring}. All models are trained from scratch with a sequence length of 1024 and an effective batch size of 4096 ($\approx 4.2$M tokens per step), identical across optimizers. Following the Chinchilla rule~\citep{hoffmann2022training} of $\approx 20$ training tokens per parameter, we train for 1750 steps ($7.3$B tokens) at 350M and 5000 steps ($21.0$B tokens) at 1.3B. We use a cosine schedule with $10\%$ linear warmup decaying to $0.1\times$ the peak learning rate, weight decay $0.1$, gradient clipping at $1.0$, $\beta_1=\beta_2=0.95$, and bfloat16 parameters and gradients. Training is carried out on nodes of up to four NVIDIA H100 80GB GPUs. 
\vspace{-0.2cm}
\paragraph{Baselines and tuning.}
We compare against AdamW~\citep{kingma2014adam}, SOAP~\citep{vyas2024soap}, Muon~\citep{jordan2024muon}, and ALSO~\citep{romashov2026twosided}. At 350M we sweep three learning rates per method and report the best run in Table~\ref{tab:perplexity}. Due to the prohibitive cost of a full sweep at 1.3B, we performed a shortened sweep and resumed only the best configurations to 5,000 steps.
\ourmethod, \ourqmethod and ALSO all use rank $r=256$ at 350M and $r=512$ at 1.3B, and all preconditioned methods refresh their preconditioner every $T=10$ steps. 
\input{tables/perplexities}
\vspace{-0.2cm}
\paragraph{Results.} Table~\ref{tab:perplexity} summarizes the final results, demonstrating that \ourmethod achieves competitive performance compared to SOAP, at a fraction of its cost. \ourqmethod cuts the optimizer state memory consumption by more than 50\% compared to Muon, and still delivers results on par with the full-rank methods. Fig.~\ref{fig:val-perplexity} plots validation perplexity throughout training for LLaMA-1.3B, where \ourmethod follows SOAP closely and presents on-par results.
\vspace{-0.2cm}
\paragraph{Efficiency.} Clean's memory advantage grows with scale (Figs.~\ref{fig:13b-mem} and \ref{fig:mem-modelsize}) because its memory complexity scales linearly with matrix dimensions, at a fixed rank $r$, whereas every baseline carries at least one quadratic term (Table~\ref{tab:optimizer_memory}). Furthermore, Fig.~\ref{fig:13b-mem} demonstrates a reduction in per-parameter memory cost as model size increases. See Appendix~\ref{app:mem-discussion} for more details.
\input{images/convergence_efficiency}
To evaluate sample efficiency and convergence, Fig.~\ref{fig:convergence} compares the methods across time and token budgets rather than focusing solely on the final loss. 
\ourmethod reaches AdamW's final performance in 64 hours instead of 81, while simultaneously reducing the optimizer memory footprint. It also only consumes 77\% of the overall token budget, with \ourqmethod requiring 87\%.
\vspace{-0.2cm}
\paragraph{Ablations.} To better investigate the sensitivity of \ourmethod, we performed ablation studies on the LLaMA-350M pre-training, focusing on two key settings: the projection rank $r$ and the preconditioner update interval $T$. In each experiment, we vary one parameter while keeping the other fixed at our default configuration ($T=10$ and $r=256$). First, Fig.~\ref{fig:rank-ablation} shows that the validation perplexity improves as the rank increases up to 256. While heavily compressed ranks ($r = 32, 64$) fail to catch up to the baselines, setting $r = 128$ achieves competitive results, and increasing the rank further to $r = 512$ yields no additional gains. Second, Fig.~\ref{fig:freq-ablation} demonstrates that $T = 10$ provides an optimal balance; while refreshing the preconditioners either too frequently ($T = 1$) or at large intervals ($T = 100$) negatively impacts the final performance.

%% file: images/val_perplexity.tex
\begin{wrapfigure}{r}{0.5\textwidth}
\vspace{-10pt}
    \centering
    \includegraphics[width=\linewidth]{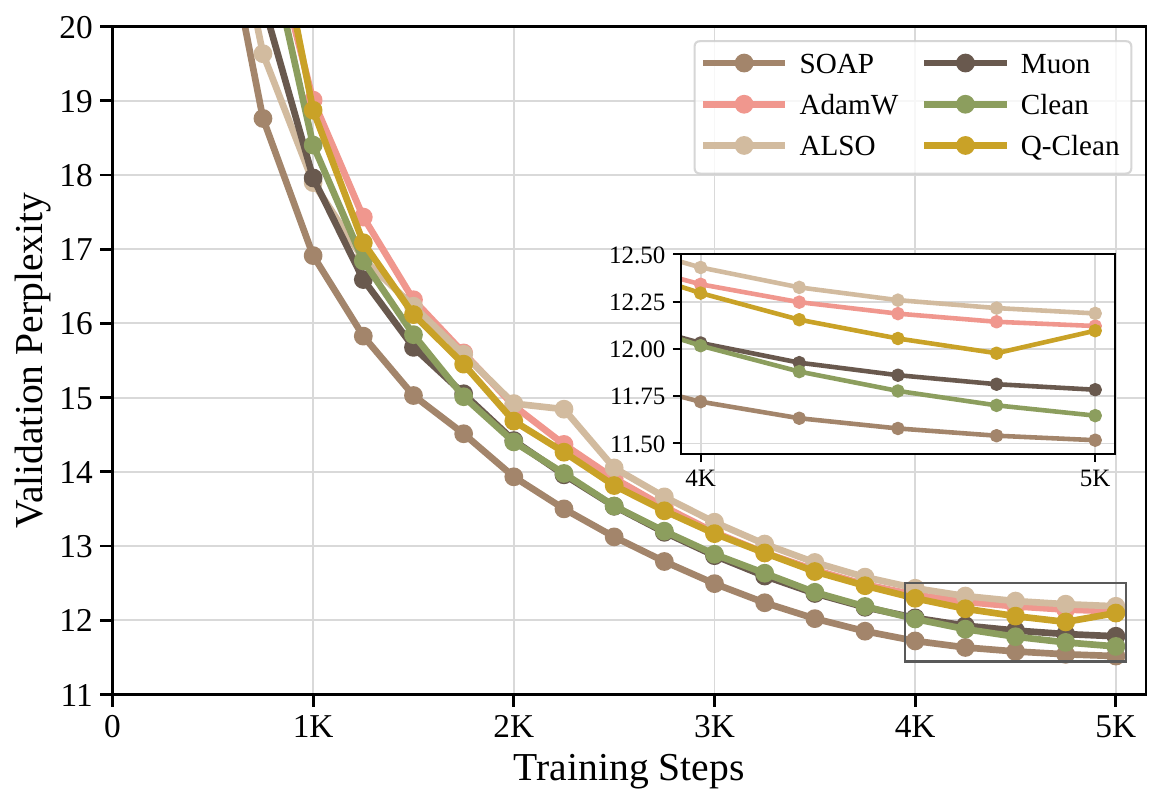}
    \caption{\small Validation perplexity recorded during the training of a LLaMA-1.3B architecture across various optimization methods.}
    \label{fig:val-perplexity}
    \vspace{-10pt}
\end{wrapfigure}

%% file: tables/perplexities.tex
\begin{table*}[t]
\small
\centering
\caption{\small Final validation perplexity ($\downarrow$) and optimizer memory when pre-training LLaMA models of different sizes. \ourmethod achieves competitive perplexity compared to full SOAP, while using less memory than AdamW. \ourmethod, \ourqmethod and ALSO use rank $r=256$ at 350M and $r=512$ at 1.3B. Excluding SOAP, the best results are shown in \textbf{bold} and the second best are \underline{underlined}.}
\label{tab:perplexity}
\renewcommand{\arraystretch}{1.15}
\resizebox{\linewidth}{!}{%
\begin{tabular}{l|ccc|ccc}
\toprule
\multirow{2}{*}{Methods}
  & \multicolumn{3}{c|}{LLaMA-350M}
  & \multicolumn{3}{c}{LLaMA-1.3B} \\
  & Perplexity & Opt.\ Mem. & Peak Mem.
  & Perplexity & Opt.\ Mem. & Peak Mem. \\
\midrule
AdamW~\citep{kingma2014adam}         & 19.63 & 1.40G & 47.93G        & 12.12 & 5.11G  & 54.13G       \\
SOAP~\citep{vyas2024soap}          & 15.12 & 7.64G & 54.25G      & 11.52 & 13.62G & 79.02G       \\
Muon~\citep{jordan2024muon}          &  16.97     &  0.83G     &  47.36G       & \underline{11.78} & 2.80G  & 51.83G       \\
ALSO~\citep{romashov2026twosided}          & \textbf{15.36} & 1.30G & 47.88G        & 12.19 & 4.71G  & 53.82G       \\
\midrule
\rowcolor{ourcolor}
\ourmethod (Ours)   & \underline{15.87} & 1.25G &  47.68G       & \textbf{11.65} & 4.48G  & 53.44G       \\
\rowcolor{ourqcolor}
{\ourqmethod (Ours)} & 16.44 & 0.38G & 46.82G       & 12.10 & 1.28G  &  50.24G      \\
\bottomrule
\end{tabular}
}
\end{table*}

%% file: images/convergence_efficiency.tex
\begin{figure*}[t]
    \centering
    \begin{subfigure}{0.497\linewidth}
    \includegraphics[width=\linewidth]{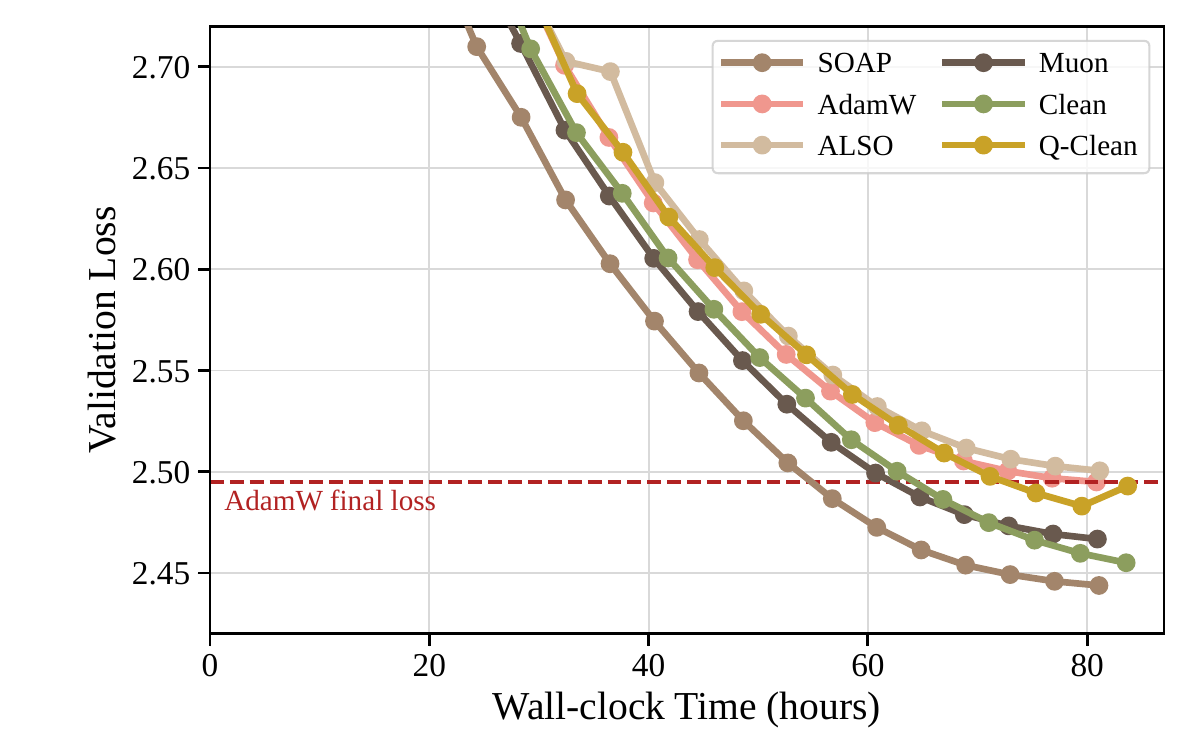}
    \caption{\small Validation loss against wall-clock time.}
    \label{fig:loss-vs-time}
    \end{subfigure}
    \hfill%
    \begin{subfigure}{0.497\linewidth}
    \includegraphics[width=\linewidth]{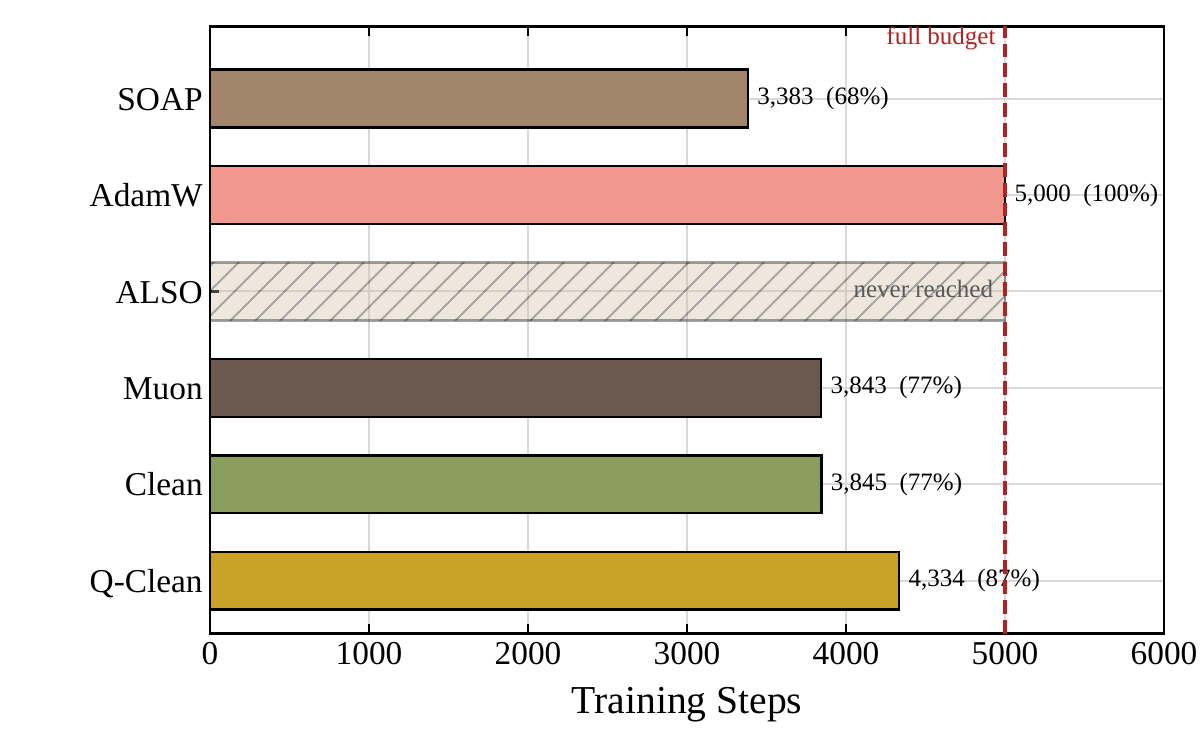}
    \caption{\small Steps to reach AdamW's final loss.}
    \label{fig:steps-to-target}
    \end{subfigure}
    \caption{\small Convergence efficiency on pre-training LLaMA-1.3B, all runs on four H100 GPUs with the same batch size, schedule and token budget. \textbf{(a) Wall-clock:} \ourmethod reaches AdamW's final validation loss in $64$ of the $81$ hours of AdamW requirements, demonstrating 26\% speedup. \textbf{(b) Token efficiency:} \ourmethod reaches that loss in $77\%$ of the training budget and \ourqmethod in $87\%$.}
    \label{fig:convergence}
    \vspace{-15pt}
\end{figure*}

%% file: sections/conclusion.tex
\section{Conclusion}
\label{sec:conclusion}

We introduce \ourmethod, a memory-efficient second-order optimizer that reduces the memory complexity of two-sided preconditioning from quadratic to linear. By leveraging randomized Nyström sketching, \ourmethod accurately approximates left and right preconditioners, while uniquely reintegrating off-subspace residual blocks to preserve essential training signals. We also develop \ourqmethod, a low-precision variant that aggressively compresses optimizer states.

Empirically, both \ourmethod and \ourqmethod achieve validation perplexities that are highly competitive with full-rank methods like SOAP, yet they operate with a smaller memory footprint than standard AdamW. Crucially, this aggressive memory reduction uniquely enables the pre-training of a 13B parameter architecture on a single 80GB GPU, and increases the convergence speed by 26\% compared to AdamW. We are optimistic that these linear-scaling curvature approximations will encourage broader exploration into accessible, memory-efficient optimization frameworks.

%% file: sections/appendix.tex
\newpage
\appendix
\section{Nystr{\"o}m Algorithm}\label{app:nystrom}
\begin{algorithm}[h!]    
	\centering
	\caption{Randomized Nystr{\"o}m Approximation~(RandNystromApprox)}
    \label{alg:nystrom-stable-appendix}
	\begin{algorithmic}[1]
		\STATE{\bfseries Input:}{ Matrix $A\in\RR^{m\times m}$ (PSD), rank size $r\leq m$, $\Omega\in\RR^{m\times r}$}.
        \\ \hrulefill
        \indent
		\STATE $\Omega\sim\mathcal{N}(0,I)\in\RR^{m\times r}$ \hfill\COMMENT{random Gaussian matrix}
	\STATE{$\Omega\gets\text{QR}(\Omega, 0)$}        \hfill\COMMENT{thin QR decomposition}
    \STATE $Y\gets A\Omega$
		\STATE{$\nu\gets\text{eps}(\text{norm}(Y, 2))$} \hfill\COMMENT{compute shift}
		\STATE{$Y_{\nu}\gets Y + \nu \Omega$} \hfill\COMMENT{add shift for stability}
        \STATE $B\gets\Omega^\ts Y_{\nu}$
		\STATE{$C\gets\text{CHOL}((B+B^\ts)/2)$} \hfill\COMMENT{Force symmetry and Cholesky decomposition}
		\STATE{$B\gets Y_{\nu} / C$} \hfill\COMMENT{triangular solve}
		\STATE{$[U, \Sigma, \sim] = \text{SVD}(B, 0)$} \hfill\COMMENT{thin SVD}
		\STATE{$\hat \Lambda\gets \text{max}\{0, \Sigma^2 - \nu I\}$} \hfill\COMMENT{remove shift, compute eigs}
		\STATE{\bfseries Output:}~$[U,\hat{\Lambda}]$
	\end{algorithmic}\label{alg:nystrom}
\end{algorithm}
Nyström approximations are well-suited for optimizer states due to their computational efficiency and solid theoretical foundations \citep{gittens2016revisiting}. While a direct implementation using Eqn.~\eqref{eqn:nystrom} suffers from numerical instability, Alg.~\ref{alg:nystrom} introduces a robust, numerically stable alternative for generating randomized Nyström approximations, as proposed in \cite{tropp2017fixed} and \cite{frangella2023randomized}. 

In this paper, rather than sampling a purely random matrix at each step, we warm-start the sketch using the eigenvectors from the previous step ($U_{prev}$), augmented with $p$ columns of fresh Gaussian noise to continuously explore new gradient directions. To further improve approximation accuracy, we perform $q$ steps of power iteration with periodic orthogonalization. Finally, we apply sign alignment to the resulting eigenvectors to prevent abrupt subspace flips, ensuring temporal consistency across consecutive optimization steps. The pseudo-code of this updated method is presented in Alg.~\ref{alg:warmup-nystrom}.
\input{algorithms/warmstart-nystrom}

\section{Quantization Details}
\label{app:quantization}
In this section, we explain the \ourqmethod optimizer in more detail. 
\subsection{Block-wise $\mathbf{n}$-bit Quantization}\label{sub:blockwise-quant}
Block-wise quantization was originally introduced by~\citep{dettmers20218} for compressing optimizer states during training, and was later adapted by~\citep{dettmers2023qlora} to frozen base-model weights for memory-efficient LLM fine-tuning. We adopt the same block-wise principle but use a uniform affine quantizer, with an \textit{asymmetric} formulation (scale and zero-point) for most matrices and a \emph{symmetric} formulation for the second moment, as detailed below.

The block-wise scheme partitions a matrix into contiguous blocks and computes an independent scale $s$ (and, in the asymmetric case, a zero-point $z$) for each block. The motivation is robustness to outliers: with a single per matrix scale, even one anomalously large value forces every quantization step to be coarse, mapping most typical values to only a handful of integer codes and losing most of the resolution~\citep{dettmers2022gpt3}. A per-block scale isolates this effect: only the block containing the outlier loses precision, while every other block keeps a fine-grained scale fitted to its own local data.
\paragraph{Asymmetric Quantization (Default).}
Given a matrix $X \in \RR^{d_1 \times d_2}$, we flatten and partition it into $N = \lceil d_1 d_2 / B \rceil$ contiguous blocks $\{X^{(i)}\}_{i=1}^N$ of size at most $B$ (the last block may be smaller). Each block is quantized independently with a per-block scale $s_i$ and zero-point $z_i$ computed from the block's full range:
\begin{equation}
\small
 s_i = \frac{\max(X^{(i)}) - \min(X^{(i)})}{2^n - 1}, \quad
z_i = -\round{\min(X^{(i)})/s_i}  
\end{equation}
Then, the quantization and dequantization maps are
\begin{equation}
\small
\begin{split}
X_q^{(i)} &= \text{Quant}_n^{\text{asym}}(X^{(i)}, s_i, z_i) = \text{clamp}\!\left(\round{\tfrac{X^{(i)}}{s_i}} + z_i,\; 0,\; 2^n - 1\right),\\
\widehat{X}^{(i)} &= \text{Dequant}_n^{\text{asym}}(X_q^{(i)}, s_i, z_i)= s_i\bigl(X_q^{(i)} - z_i\bigr) \;\approx\; X^{(i)}
\end{split}
\end{equation}
where $\round{\cdot}$ denotes round-to-nearest and $n$ is the number of bits. We apply this affine form with the stochastic-rounding explained in section~\ref{sub:quant-sr} to the projection matrices ($U_{t, L}$ and $U_{t, R}$) and to all first-moment matrices ($M_t$, $M_{t, A}$, and $M_{t, B}$). 
However, we do not use this quantization for the second-moment matrices, because their strictly non-negative, highly skewed distributions and large dynamic ranges make linear methods suboptimal. Instead, we apply a non-linear symmetric quantization approach for the main second moment. The second moments of blocks \texttt{A} and \texttt{B} remain in full precision; as simple size-$r$ vectors, their memory footprint is already negligible and quantizing them offers no meaningful savings.

\paragraph{Non-linear Quantization for the Second-moment.}
Because the second moment statistics are strictly non-negative ($X^{(i)}\geq 0$) and exhibit a highly skewed distribution with a large dynamic range, standard linear quantization allocates insufficient resolution to the majority of values concentrated near zero. To address this, we utilize the non-linear quantization scheme from \texttt{bitsandbytes} \citep{dettmers20218}.

Rather than spacing bins equally, this method employs a fixed non-linear codebook $\mathcal{C}$ consisting of 256 quantization levels (for 8-bit representation). These levels are dynamically spaced to provide high precision for small-magnitude values while remaining capable of capturing large outliers. For a given block, we first compute the maximum value to serve as the block-wise scaling factor:
\begin{equation}
\small
s_i = \max(X^{(i)})
\end{equation}
The elements of the block are normalized by $s_i$ to lie in the range $[0, 1]$. Each normalized value is then mapped to the nearest entry in the codebook $\mathcal{C}$. The quantization and dequantization mappings are formalized as:
\begin{equation}
\small
\begin{split}
X_q^{(i)} &= \text{Quant}_{\text{8-bit}}^{\text{NL}}(X^{(i)}, s_i) = \argmin_{c \in \mathcal{C}} \left| \frac{X^{(i)}}{s_i} - c \right|, \\
\widehat{X}^{(i)} &= \text{Dequant}_{\text{8-bit}}^{\text{NL}}(X_q^{(i)}, s_i) = s_i \cdot X_q^{(i)},
\end{split}
\end{equation}
where $X_q^{(i)}$ is stored in memory as an 8-bit integer corresponding to the index of the matched codebook entry $c$. This formulation eliminates the need for a zero-point, requiring only the per-block scale $s_i$ as metadata, while perfectly matching the non-linear, non-negative distribution of the optimizer's second moments.
\subsection{Maintaining High-precision Training with Stochastic Rounding Accumulated Quantization}\label{sub:quant-sr}
To reduce memory footprint, we store optimizer states as matrices in low precision (INT8) throughout the forward and backward passes. Doing so, however, introduces a subtle but fatal failure mode: under round-to-nearest (RTN), if the update magnitude is not large enough, the effect will be mitigated by RTN. 
Following Q-GaLore~\citep{zhang2024q}, we 
apply stochastic rounding (SR)~\citep{von1947numerical,gupta2015deep} in our asymmetric quantization. A floating-point value $X$ lying between
two adjacent quantization levels $\lfloor X \rfloor$ and $\lceil X \rceil$ is
rounded \emph{up} to $\lceil X \rceil$ with probability $X - \lfloor X \rfloor$
and rounded \emph{down} to $\lfloor X \rfloor$ otherwise. This rule is unbiased,
i.e.\ $\EE(X_q) = X$. More concretely, the two rounding schemes, namely rounding-to-nearest~(RTN) and stochastic rounding, are defined as follows, respectively:
\begin{align}
\small
    \round{X} &=
    \begin{cases}
            \lfloor X \rfloor
                & \text{if } \lfloor X \rfloor \leq X < \lfloor X \rfloor + \tfrac{\epsilon}{2}, \\[2pt]
            \lfloor X \rfloor + \epsilon
                & \text{if } \lfloor X \rfloor + \tfrac{\epsilon}{2} \leq X \leq \lfloor X \rfloor + \epsilon,
        \end{cases}
    \end{align}
\begin{align}
\small
\text{round}(X) =
        \begin{cases}
            \lfloor X \rfloor
                & \text{with probability } \lceil X \rceil - X, \\[2pt]
            \lceil X \rceil
                & \text{with probability } X - \lfloor X \rfloor,
        \end{cases}
\end{align}
where $\epsilon = 2^{-\mathrm{FL}}$ and $\mathrm{FL}$ denotes the length of the
fractional part.

\subsection{Quantizer Choice}
We tracked every matrix family $W$ (weights), $G$ (gradients), $M$ and $V$ (Adam moments), $L$ and $R$ (SOAP's accumulated covariance matrices), and $Q_L$ and $Q_R$ (eigenbasis of $L$ and $R$) at every layer of the LLaMA-350M model while optimizing via SOAP (see Alg.~\ref{alg:soap}). For each family, we then pool all captured matrices and report empirical densities of three quantities: scale ($\log_{10}\max|x|$), dynamic range $\left(\log_{10}^{\bigl(\max|x|/\mathrm{median}|x|\bigr)}\right)$, and the non-excess kurtosis ($\log_{10}\kappa$), with $\kappa=\mathbb{E}[(x-\mu)^4]/\sigma^4$.

  As shown in Figure~\ref{fig:distributions}, we evaluate the statistics using two perspectives based on the quantization strategy. The per-matrix view computes a single scalar for an entire matrix, highlighting differences between matrices; this applies when sharing one quantization scale across a matrix. Conversely, the per-block view divides matrices into blocks, computing and pooling statistics for each block individually.

 Together, these views motivate our quantization strategy. In the per-matrix view, overall matrices ($V$, $L$, $R$, etc.) exhibit extreme distributional probabilities; a global linear scale stretches to cover these heavy tails, wasting resolution and causing high error. Moving to a block-wise approach enhances the tiles by isolating these extreme values, which lightens the tails across all families. For most matrices ($W$, $G$, $M$, $L$, $R$, $Q_L$, $Q_R$), this shift makes them suitable for asymmetric and linear quantization. However, even after block-wise quantization, the second-moment matrix ($V$) still shows heavy tails, and linear quantization leads to divergence on $V$, whereas it succeeds for the other matrices. Therefore, we use a symmetric and non-linear quantizer for $V$ and retain linear block-wise codes for the rest in our low-rank proposed method.
 \input{images/distibutions}

\section{SOAP}
\label{app:soap}
Algorithm~\ref{alg:soap} presents the pseudocode for the SOAP optimizer~\citep{vyas2024soap}. As shown, the full $L$ and $R$ covariance matrices are constructed and stored, while the Adam optimizer is applied to the gradient projected onto their eigenbasis. Although SOAP demonstrates accelerated convergence compared to other standard optimizers, it introduces prohibitive memory requirements by simultaneously maintaining both the covariance matrices and the full Adam moments.
\input{algorithms/soap}

\section{Linear Memory Discussion}
\label{app:mem-discussion}
The linear memory complexity of \ourmethod and \ourqmethod with respect to model dimensions makes them exceptionally scalable. 

As demonstrated in Figure~\ref{fig:13b-mem} (left), \ourmethod and \ourqmethod are the only optimizers that allow pre-training a LLaMA-13B architecture on a single 80 GB H100 GPU. Notably, the other low-rank second-order baseline, ALSO~\citep{romashov2026twosided}, requires more than 80 GB of memory even at rank $r = 32$. This is due to the fact that \ourmethod keeps no matrix with dimensions of order $\mathcal{O}(mn)$ or $\mathcal{O}(m^2+n^2)$, while all other optimizers have at least one $\mathcal{O}(mn)$ component.

Furthermore, Figure~\ref{fig:13b-mem} (right) illustrates the cost per parameter for different optimizers. Remarkably, our methods are the only ones that reduce this per-parameter cost compared to the baselines. To ensure a realistic setup, we scale the rank proportionally with the model dimension, setting $r = d/8$, where $d=\min(m,n)$ is the smaller dimension of each weight matrix. This trend is highly promising in terms of democratizing training of much larger foundation models on more accessible hardware.
\input{images/memory_ablation_13B.tex}

Additionally, Figure~\ref{fig:mem-modelsize} illustrates the effect of varying ranks across different model sizes for \ourmethod, \ourqmethod, and the low-rank second-order baseline, ALSO~\citep{romashov2026twosided}. Figure~\ref{fig:mem-modelsize} (left) demonstrates that for a fixed rank of $r = 32$, the memory savings gap between our methods and the baseline increases as the model size scales. This improvement stems from the omission of quadratic memory terms, allowing our methods to operate at a significantly lower cost. Furthermore, Figure~\ref{fig:mem-modelsize} (right) compares how the memory consumption of ALSO scales relative to our methods across different ranks and model dimensions. As expected, the memory efficiency gap widens as the model size increases.
\input{images/complexity_comparison.tex}

\section{Ablating Hyperparameters}
Here we provide the supporting figures for the hyperparameter ablation studies discussed in section~\ref{sec:experiment}. Fig. \ref{fig:rank-ablation} presents the learning curves and validation perplexities for varying projection ranks ($r$) on the LLaMA-350M pre-training task. Figure \ref{fig:freq-ablation} details the performance across different preconditioner update intervals ($T$). These figures illustrate the full training trajectories, confirming the optimal configurations of $r = 256$ and $T = 10$ established in our primary analysis.
\input{images/ablations}

\section{Hyperparameters}
\label{app:hyperparameters}
\input{tables/hyperparameters}

%% file: algorithms/warmstart-nystrom.tex
\begin{algorithm}[h!]
\caption{Warm-started Randomized Nystr{\"o}m with Subspace Iteration}
\begin{algorithmic}[1]
\REQUIRE $A\in\RR^{m\times m}$ (PSD),
       target rank $r$, oversampling $p$, power-iterations passes $q$, warm start $U_{\text{prev}}\in\RR^{m\times r}$ (or $\emptyset$)
\vspace{5pt}
\IF{$U_{\text{prev}}\neq\emptyset$}
   \STATE $\Psi\sim\mathcal N(0,I)\in\RR^{m\times p}$;\quad
          $\Omega\gets[\,U_{\text{prev}},\,\Psi\,]$  \hfill\COMMENT{warm start + fresh exploration block}
\ELSE
   \STATE $\Omega\sim\mathcal N(0,I)\in\RR^{m\times(r+p)}$
\ENDIF
\STATE $\Omega\gets\text{QR}(\Omega,0)$
\FOR{$j=1$ {\bfseries to} $q$}
   \STATE $Y\gets A\Omega$
   \STATE $\Omega\gets\text{QR}(Y,0)$ \hfill\COMMENT{re-orthonormalize \emph{every} pass}
\ENDFOR
\STATE $Y\gets A\Omega$ \hfill\COMMENT{final sketch must match the final $\Omega$}
\STATE $\nu\gets\text{eps}(\text{norm}(Y,2))$
\STATE $Y_\nu\gets Y+\nu\Omega$
\STATE $B\gets\Omega^\ts Y_\nu$
\STATE $C\gets\text{CHOL}((B+B^\ts)/2)$
\STATE $[U,\Sigma,\sim]\gets\text{SVD}(Y_\nu/C,0)$
\STATE $\hat\Lambda\gets\max\{0,\Sigma^2-\nu I\}$
\STATE $U\gets U_{:,1:r}$,\quad $\hat\Lambda\gets\hat\Lambda_{1:r,1:r}$ \hfill\COMMENT{discard oversampling}
\STATE $s_i\gets\text{sign}(u_i^\ts u_{\text{prev},i})$;\quad $U_{:,i}\gets s_i U_{:,i}$ \hfill\COMMENT{sign alignment}
\STATE {\bfseries Output:} $[U,\hat\Lambda]$
\end{algorithmic}\label{alg:warmup-nystrom}
\end{algorithm}

%% file: images/distibutions.tex
\begin{figure*}[th!]

 \scriptsize
    \centering
    \begin{subfigure}{0.325\textwidth}
        \centering
        \includegraphics[width=\linewidth]{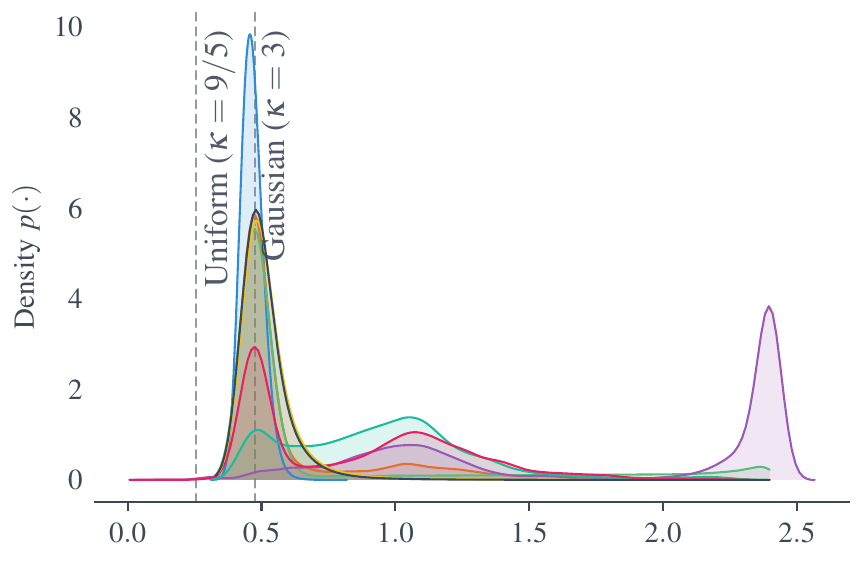}
        \caption{\scriptsize $\log_{10}^{\kappa}$ per block ($\kappa = \mathbb{E}[(x-\mu)^4]/\sigma^4$).}
    \end{subfigure}
    \begin{subfigure}{0.325\textwidth}
        \centering
        \includegraphics[width=\linewidth]{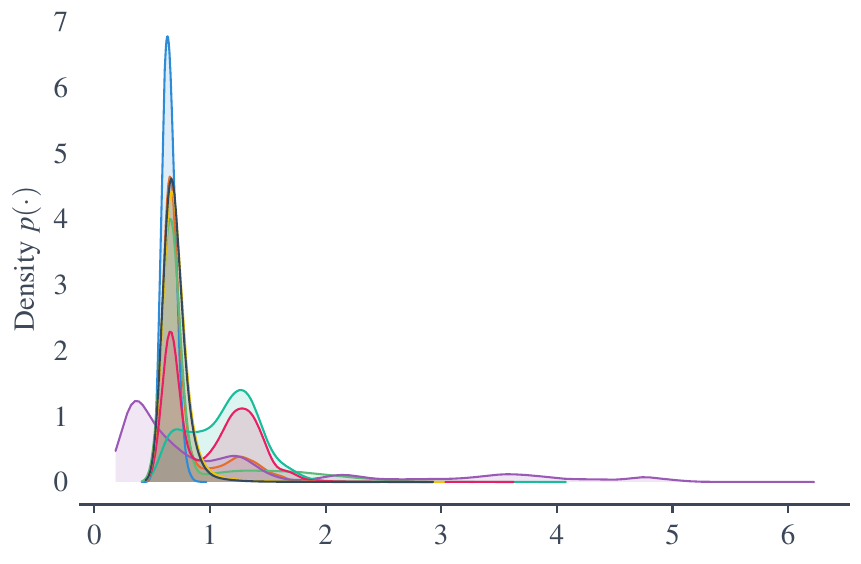}
        \caption{\scriptsize $\log_{10}^{max(|x|)/median(|x|)}$ per block.}
    \end{subfigure}
    \begin{subfigure}{0.325\textwidth}
        \centering
        \includegraphics[width=\linewidth]{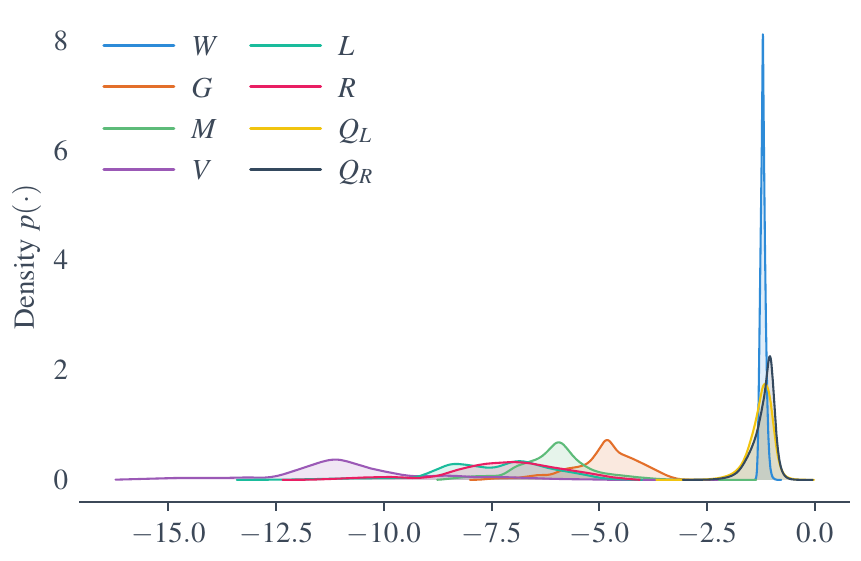}
        \caption{\scriptsize $\log_{10}^{\max(|x|)}$ per block.}
    \end{subfigure}

    \begin{subfigure}{0.325\textwidth}
        \centering
        \includegraphics[width=\linewidth]{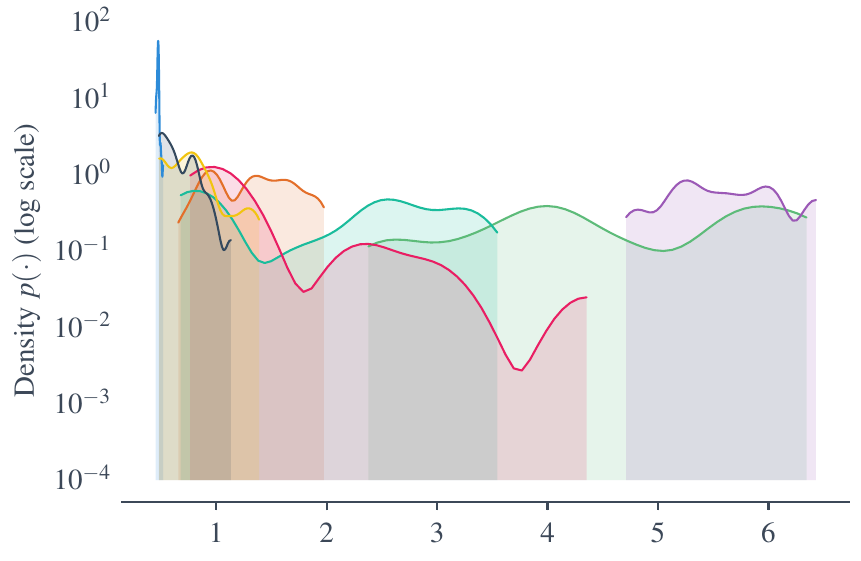}
        \caption{\scriptsize $\log_{10}^{\kappa}$ per matrix ($\kappa = \mathbb{E}[(x-\mu)^4]/\sigma^4$).}
    \end{subfigure}
    \begin{subfigure}{0.325\textwidth}
        \centering
        \includegraphics[width=\linewidth]{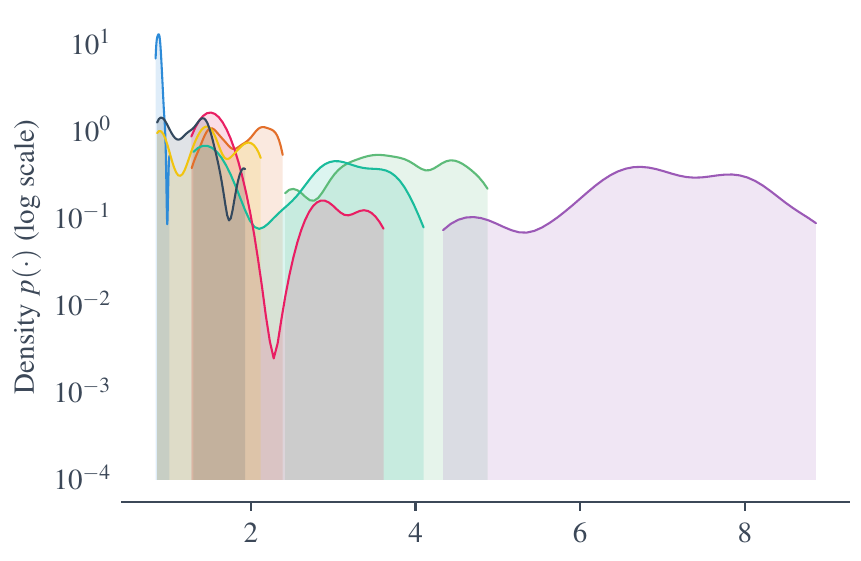}
        \caption{\scriptsize $\log_{10}^{max(|x|)/median(|x|)}$ per matrix.}
    \end{subfigure}
    \begin{subfigure}{0.325\textwidth}
        \centering
        \includegraphics[width=\linewidth]{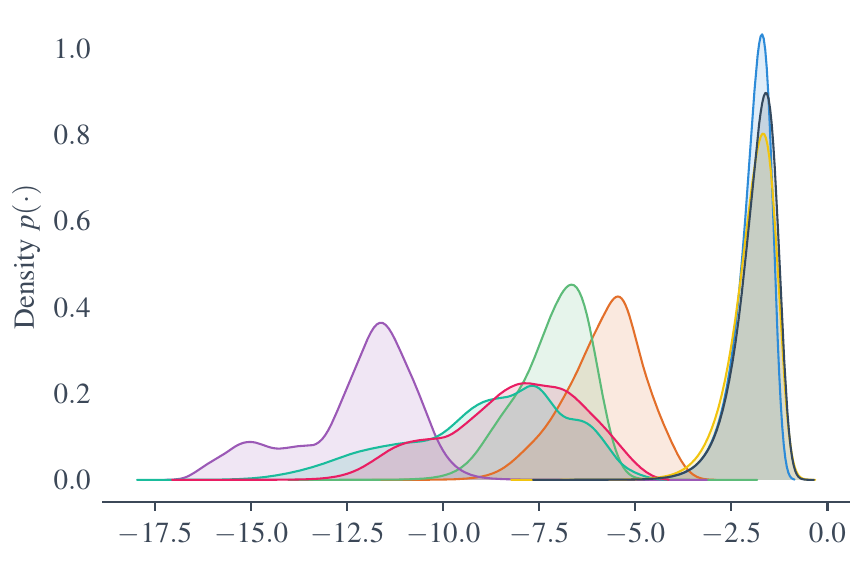}
        \caption{\scriptsize $\log_{10}^{\max(|x|)}$ per matrix  .}
    \end{subfigure}
    \caption{The plots compare non-excess kurtosis, dynamic range, and scale under two quantization granularities: per-block (a–c) and per-matrix (d–f). The shift from a per-matrix to a per-block approach significantly reduces the heavy-tailed distributions for most matrices, rendering them suitable for linear quantization. However, the second moment matrix ($V$, in purple) remains a prominent outlier, exhibiting persistent heavy tails even in the per-block view, which motivates the use of a non-linear quantizer specifically for $V$.}
    \label{fig:distributions}
\end{figure*}

%% file: algorithms/soap.tex
\begin{algorithm}[H]
	\begin{algorithmic}[1]
		\STATE Sample batch $B_t$.
		\STATE $G \in \mathbb{R}^{m \times n} \gets -\nabla_W \phi_{B_t}(W_t)$
		\STATE $G' \gets Q_L^T G Q_R$
		\STATE $M \gets \beta_1 M + (1-\beta_1) G$
		\STATE $M' \gets Q_L^T M Q_R$
		\STATE \COMMENT{Now we  ``run'' Adam on $G'$}
		\STATE $V \gets \beta_2 V + (1-\beta_2) (G' \odot G')$ \COMMENT{Elementwise multiplication}
		\STATE $N' \gets \frac{M'}{\sqrt{\hat{V}_t} + \epsilon}$ \COMMENT{Elementwise division and square root}
		\STATE \COMMENT{Now that we have preconditioned by Adam in the rotated space, we go back to the original space.}
		\STATE $N \gets Q_L N' Q_R^T$ 
		\STATE $W \gets W - \eta N$
		\STATE \COMMENT{End of gradient step, we now update $L$ and $R$ and possibly also $Q_L$ and $Q_R$. }
		\STATE $L \gets \beta_2  L + (1-\beta_2) GG^T$
		\STATE $R \gets \beta_2  R + (1-\beta_2) G^TG$
		\IF{t \% f == 0}
		\STATE $Q_L \gets \texttt{Eigenvectors}(L, Q_L)$
		\STATE $Q_R \gets \texttt{Eigenvectors}(R, Q_R)$
		\ENDIF
\end{algorithmic}
	\caption{SOAP}
	\label{alg:soap}
\end{algorithm}

%% file: images/memory_ablation_13B.tex
\begin{figure}[t]
    \centering
    \includegraphics[width=\linewidth]{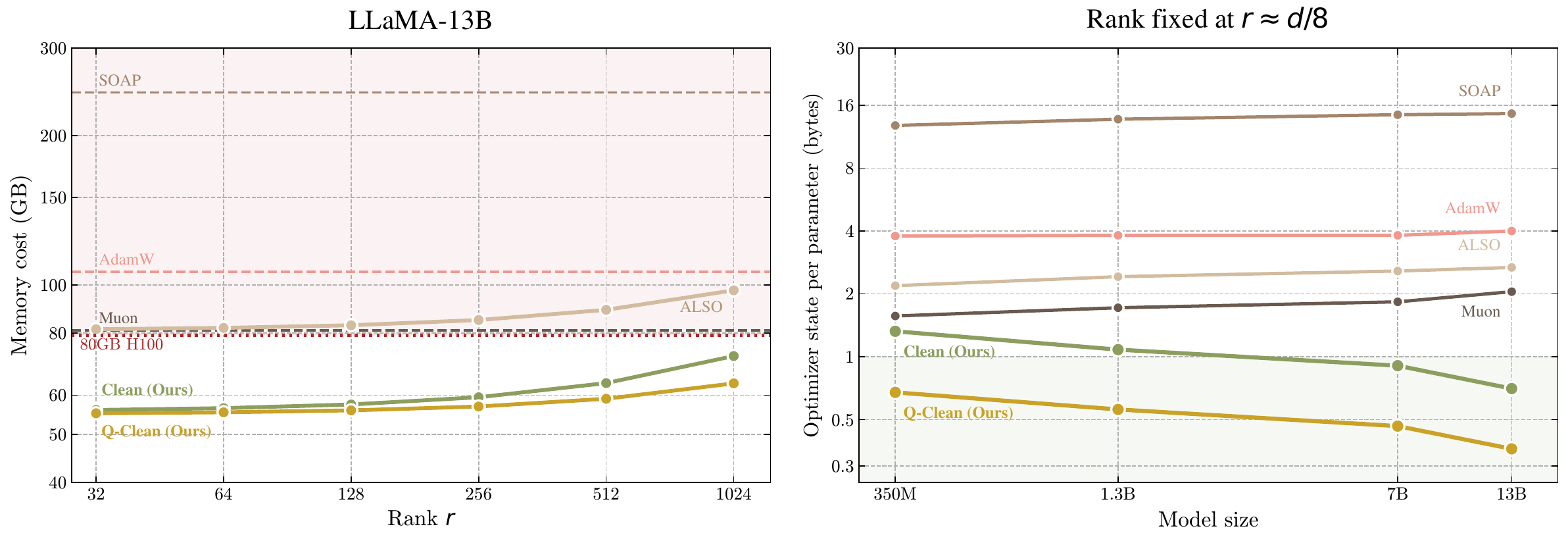}
    \caption{\small \textbf{Left:} total memory for pre-training LLaMA-13B on a single 80\,GB H100 as a function of rank; the shaded region is infeasible for all of our baselines at any rank. \textbf{Right:} optimizer state per parameter across model sizes at $r\approx d/8$. Every baseline carries an $mn$ term (Table~\ref{tab:optimizer_memory}) and so stays flat or rises; \ourmethod{} and \ourqmethod{} scale as $r(m+n)$ and are the only methods whose per-parameter cost \emph{decreases} with model size.}
    \label{fig:13b-mem}
\end{figure}

%% file: images/complexity_comparison.tex
\begin{figure}[h!]
  \centering
  \includegraphics[width=\linewidth]{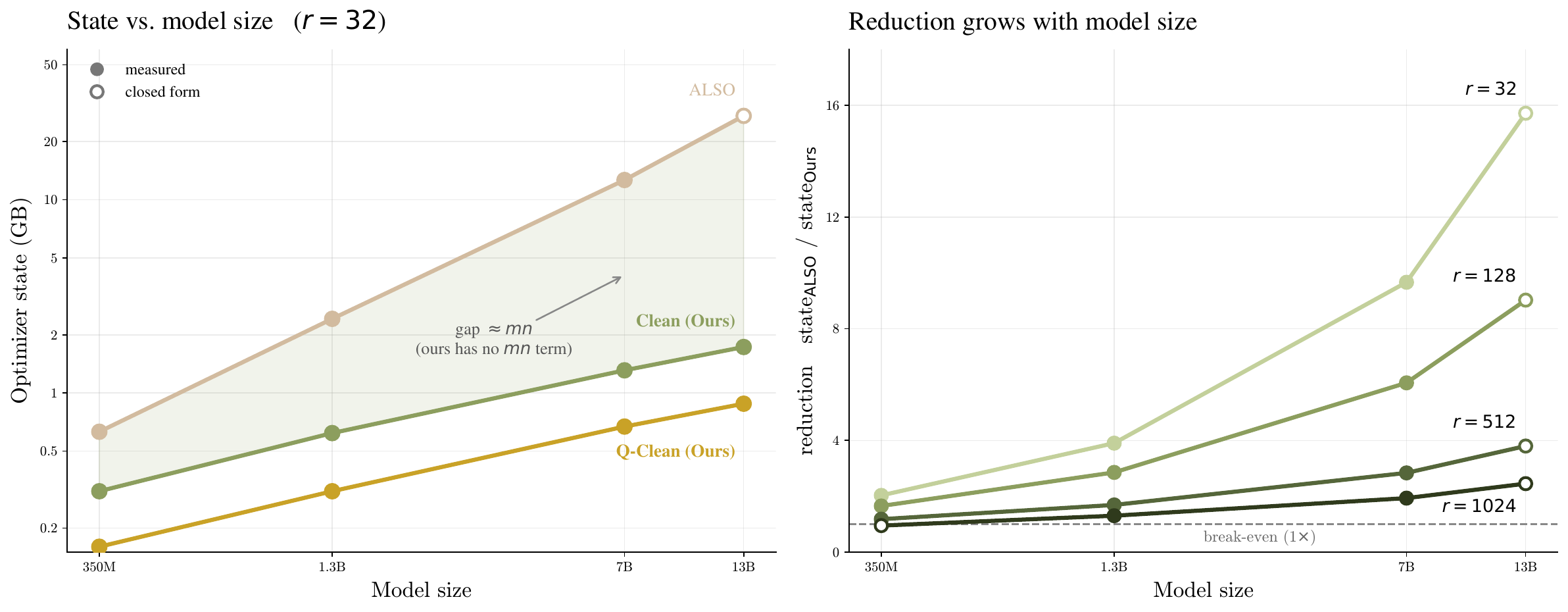}
  \caption{\textbf{Measured optimizer state on LLaMA.} Baselines carry an $mn$ term;
    ours replaces it with $r(r+2)$, so our state is smaller whenever
    $mn > r(r{+}2)$. The margin widens with scale because $mn$ grows with the model
    ($2.8$M to $70.8$M from 350M to 13B) while $r(r{+}2)$ is fixed by the rank.
    \textbf{Left:} the shaded gap is the omitted $mn$ term at $r{=}32$.
    \textbf{Right:} reduction over ALSO, $15.7\times$ at $r{=}32$ and $3.8\times$ at
    $r{=}512$ on 13B. Batch 1, seq 1024, bf16 state; filled markers measured, hollow
    closed-form where the run exceeded 80\,GB in ALSO.}
  \label{fig:mem-modelsize}
\end{figure}

%% file: images/ablations.tex
\begin{figure*}[h!]
    \centering
    \begin{subfigure}{0.497\linewidth}
    \includegraphics[width=\linewidth]{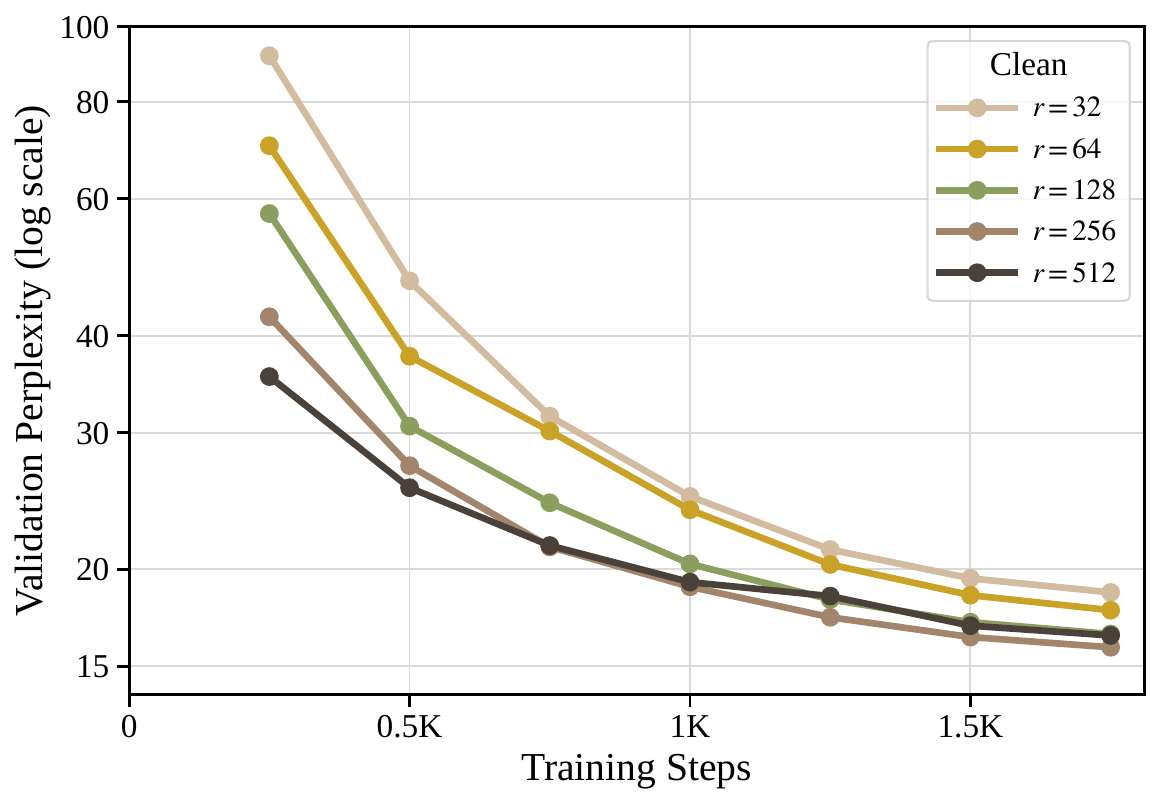}
    \caption{\small Rank $r$ in projection matrices.}
    \label{fig:rank-ablation}
    \end{subfigure}
    \hfill%
    \begin{subfigure}{0.497\linewidth}
    \includegraphics[width=\linewidth]{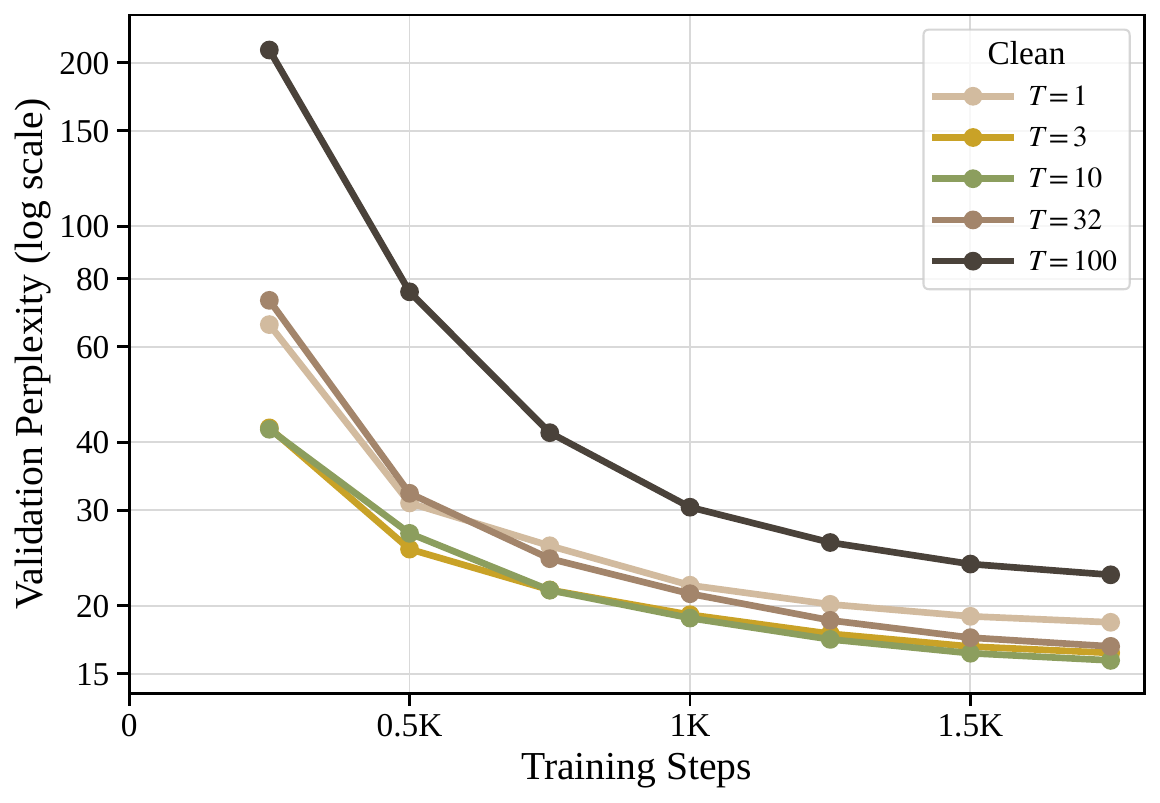}
    \caption{\small Preconditioner update frequency $T$.}
    \label{fig:freq-ablation}
    \end{subfigure}
    \caption{\small Ablations of \ourmethod on pre-training LLaMA-350M. \textbf{(a) Rank $r$:} Performance improves as rank increases up to 128 with marginal improvements afterward. \textbf{(b) Update interval $T$:} $T = 10$ is optimal; updating too often ($T = 1$) or too rarely ($T = 100$) negatively impacts performance.}
    \label{fig:ablations}
\end{figure*}

%% file: tables/hyperparameters.tex
Tables~\ref{tab:hparams_setup} and~\ref{tab:hparams_opt} list
the hyperparameters of all runs. SOAP, \ourmethod{} and
\ourqmethod{} optimize all parameters themselves: 1-D
parameters receive Adam updates with the method's own
$(\beta_1,\beta_2)$. 
\ourqmethod{} quantizes
the eigenbases and all first moments, including the
$M$, $M_{\texttt{A}}$, and $M_{\texttt{B}}$, with asymmetric per-group linear
quantization and stochastic rounding, and the second moments
with block-wise dynamic quantization; the eigenvalues and the
pooled second moments stay in original data type.

\begin{table*}[h]
\small
\centering
\caption{Model architectures and training setup shared by all optimizers. Gradient accumulation makes up the global batch. Peak memory in Table~\ref{tab:perplexity} is measured on a single GPU at this micro-batch size, without activation checkpointing.}
\label{tab:hparams_setup}
\renewcommand{\arraystretch}{1.15}
\begin{tabular}{l|cc}
\toprule
 & LLaMA-350M & LLaMA-1.3B \\
\midrule
\textbf{Architecture} & & \\
Parameters & 368M & 1.34B \\
Hidden dimension & 1024 & 2048 \\
Intermediate dimension & 2736 & 5461 \\
Attention heads & 16 & 32 \\
Layers & 24 & 24 \\
Vocabulary size &
  \multicolumn{2}{c}{32000 (LLaMA tokenizer)} \\
\midrule
\textbf{Training} & & \\
Dataset & \multicolumn{2}{c}{C4 (English)} \\
Sequence length & \multicolumn{2}{c}{1024} \\
Batch size (sequences) & \multicolumn{2}{c}{4096} \\
Micro-batch size & 32 & 16 \\
Training steps & 1750 & 5000 \\
Warmup steps & 175 & 500 \\
Learning-rate schedule &
  \multicolumn{2}{c}{linear warmup, cosine decay to
  $0.1\times$ peak} \\
Weight decay (decoupled, all parameters) &
  \multicolumn{2}{c}{0.1} \\
Gradient clipping (global norm) & \multicolumn{2}{c}{1.0} \\
Adam $\epsilon$ & \multicolumn{2}{c}{$10^{-8}$} \\
Precision & \multicolumn{2}{c}{bf16 (weights, gradients,
  activations)} \\
Evaluation &
  \multicolumn{2}{c}{C4 validation, $\approx$10M tokens} \\
\bottomrule
\end{tabular}
\end{table*}

\begin{table}[h]
\small
\centering
\caption{Learning rates swept at 350M. Three values per optimizer; the
selected value, used in Table~\ref{tab:hparams_opt} and reported in
Table~\ref{tab:perplexity}, is shown in \textbf{bold}.}
\label{tab:lr_sweep}
\renewcommand{\arraystretch}{1.15}
\begin{tabular}{l|ccc}
\toprule
Method & \multicolumn{3}{c}{Learning rate} \\
\midrule
AdamW & $7\times10^{-4}$ & $\mathbf{1.4\times10^{-3}}$ & $2.8\times10^{-3}$ \\
SOAP  & $7\times10^{-4}$ & $1.4\times10^{-3}$ & $\mathbf{2.8\times10^{-3}}$ \\
Muon  & $2\times10^{-3}$ & $6\times10^{-3}$ & $\mathbf{1.0\times10^{-2}}$ \\
ALSO  & $1.4\times10^{-3}$ & $\mathbf{4.4\times10^{-3}}$ & $1.4\times10^{-2}$ \\
\midrule
\rowcolor{ourcolor}
\ourmethod{} (Ours)  & $1.4\times10^{-3}$ & $\mathbf{4.4\times10^{-3}}$ & $1.4\times10^{-2}$ \\
\rowcolor{ourqcolor}
\ourqmethod{} (Ours) & $1.4\times10^{-3}$ & $\mathbf{4.4\times10^{-3}}$ & $1.4\times10^{-2}$ \\
\bottomrule
\end{tabular}
\end{table}

\begin{table*}[t]
\small
\centering
\caption{Optimizer-specific hyperparameters; learning rates
are peak values of the schedule in
Table~\ref{tab:hparams_setup}. 
}

\label{tab:hparams_opt}
\renewcommand{\arraystretch}{1.05}
\begin{tabular}{l|l|cc}
\toprule
Method & Hyperparameter & LLaMA-350M & LLaMA-1.3B \\
\midrule
\multirow{2}{*}{AdamW}
 & Learning rate & $1.4\times10^{-3}$ & $1.4\times10^{-3}$ \\
 & $(\beta_1, \beta_2)$ & \multicolumn{2}{c}{$(0.95, 0.95)$} \\
\midrule
\multirow{4}{*}{SOAP}
 & Learning rate & $2.8\times10^{-3}$ & $2.8\times10^{-3}$ \\
 & $(\beta_1, \beta_2)$ & \multicolumn{2}{c}{$(0.95, 0.95)$} \\
 & Shampoo $\beta$ & \multicolumn{2}{c}{0.95} \\
 & Preconditioning frequency & \multicolumn{2}{c}{10} \\
\midrule
\multirow{3}{*}{Muon}
 & Learning rate &
   $1.0\times10^{-2}$ & $1.4\times10^{-2}$ \\
 & Momentum (Nesterov) & \multicolumn{2}{c}{0.9} \\
 & Newton--Schulz iterations & \multicolumn{2}{c}{5} \\
\midrule
\multirow{6}{*}{ALSO}
 & Learning rate & $1.4\times10^{-3}$ & $4.0\times10^{-3}$ \\
 & Rank $r$ & 256 & 512 \\
 & Basis update interval $K$ & \multicolumn{2}{c}{10} \\
 & $(\beta_1, \beta_2, \beta_3)$ &
   \multicolumn{2}{c}{$(0.95, 0.95, 0.95)$} \\
 & Compensation weight $\alpha_c$ & \multicolumn{2}{c}{1.0} \\
 & Norm-growth limit $\gamma$ & \multicolumn{2}{c}{1.01} \\
\midrule
\rowcolor{ourcolor}
 & Learning rate & $4.4\times10^{-3}$ & $1.0\times10^{-3}$ \\
\rowcolor{ourcolor}
 & $(\beta_1, \beta_2)$ & \multicolumn{2}{c}{$(0.95, 0.95)$} \\
\rowcolor{ourcolor}
 & Rank $r$ & 256 & 512 \\
\rowcolor{ourcolor}
 & Oversampling $p$ & 8 & 16 \\
\rowcolor{ourcolor}
 & Power iterations & \multicolumn{2}{c}{1} \\
\rowcolor{ourcolor}
 & Preconditioning frequency $T$ & \multicolumn{2}{c}{10} \\
\rowcolor{ourcolor}
 & Shampoo $\beta$ (per refresh) & \multicolumn{2}{c}{0.95} \\
\rowcolor{ourcolor}
 & Max.\ precond.\ dimension & 4096 & 8192 \\
\rowcolor{ourcolor}
 \multirow{-9}{*}{\ourmethod{} (Ours)}
 & Norm-growth limit $\gamma$ & \multicolumn{2}{c}{1.01} \\
\midrule
\rowcolor{ourqcolor}
 & Learning rate & $4.4\times10^{-3}$ & $1.0\times10^{-3}$ \\
\rowcolor{ourqcolor}
 & Other hyperparameters &
   \multicolumn{2}{c}{as \ourmethod{}} \\
\rowcolor{ourqcolor}
 & Eigenbases $U_L, U_R$ &
   \multicolumn{2}{c}{INT8 linear, group 256, SR} \\
\rowcolor{ourqcolor}
 & First moments &
   \multicolumn{2}{c}{INT8 linear, group 256, SR} \\
\rowcolor{ourqcolor}
 & Second moments &
   \multicolumn{2}{c}{INT8 dynamic, block 256} \\
\rowcolor{ourqcolor}
\multirow{-6}{*}{\ourqmethod{} (Ours)}
 & Eigenvalues & \multicolumn{2}{c}{as \ourmethod{}} \\
\bottomrule
\end{tabular}
\end{table*}